\documentclass[letterpaper,10 pt,conference]{ieeeconf}  

\IEEEoverridecommandlockouts                              
\usepackage[latin9]{inputenc}
\usepackage{array}
\usepackage{booktabs}
\usepackage{textcomp}
\usepackage{mathrsfs}
\usepackage{multirow}
\usepackage[caption=false]{subfig}
\usepackage{amsmath,amsfonts}
\usepackage{graphicx}
\PassOptionsToPackage{normalem}{ulem}
\usepackage{ulem}
\usepackage{url}
\usepackage{balance}

\usepackage{cite}
\usepackage[pagebackref=true,breaklinks=true,letterpaper=true,colorlinks,bookmarks=true]{hyperref}

\title{\LARGE \bf FreSCo: \uline{Fre}quency-Domain \uline{S}can \uline{Co}ntext for LiDAR-based \\ Place Recognition with Translation and Rotation Invariance} 

\author{
	Yongzhi Fan$^{1,2}$, 
	Xin Du$^{1}$, 
	Lun Luo$^{1}$ 
	and Jizhong Shen$^{1*}$ 
	\thanks{$^{1}$Y. Fan, X. Du, L. Luo and J. Shen are with the 	College of 
		Information Science \& Electronic Engineering, Zhejiang University, Hangzhou 310027, China. }
	\thanks{$^{2}$Y. Fan is with the Key Laboratory of Collaborative Sensing and Autonomous Unmanned Systems of Zhejiang Province, China. }
	\thanks{$^{*}$The corresponding author. \texttt{E-mail: {jzshen@zju.edu.cn}}}
}

\begin{document}

\maketitle
\thispagestyle{empty}
\pagestyle{empty}

\begin{abstract}
Place recognition plays a crucial role in re-localization and loop closure detection tasks for robots and vehicles. 
This paper seeks a well-defined global descriptor for LiDAR-based place recognition.
Compared to local descriptors, global descriptors show remarkable performance in urban road scenes but are usually viewpoint-dependent. 
To this end, we propose a simple yet robust global descriptor dubbed FreSCo that decomposes the viewpoint difference during revisit and achieves both translation and rotation invariance by leveraging Fourier Transform and circular shift technique.
Besides, a fast two-stage pose estimation method is proposed to estimate the relative pose after place retrieval by utilizing the compact 2D point clouds extracted from the original data. 
Experiments show that FreSCo exhibited superior performance than contemporaneous methods on sequences of different scenes from multiple datasets.
Code will be publicly available at \href{https://github.com/soytony/FreSCo}{\texttt{https://github.com/soytony/FreSCo}}.
\end{abstract}


\section{Introduction}

Robots and vehicles need reliable ability of place recognition to perform re-localization on a previously obtained map or detect loop closures in large-scale simultaneous localization and mapping (SLAM).
Most approaches tackle this problem by designing descriptors for data acquired through sensors like cameras and LiDARs.
Camera-based methods often extract distinctive local features, say, points~\cite{Rublee2011,Galvez-Lopez2012,Mur-Artal2017} and line segments~\cite{Lee2014,Gomez-Ojeda2019} from the images captured at a scene.
Then they generate compact descriptors for these features and utilize the bag of words (BoW) model to implement candidate retrieval.
However, such approaches suffer from illumination, season, and viewpoint changes, making them not robust.
In contrast, LiDAR actively captures the structural information, which is less sensitive to those changes.
Thus, more attention has been drawn to LiDAR-based methods in recent years.

Existing works on LiDAR-based place recognition can be categorized into two classes, local-feature based methods~\cite{Rusu2009,Salti2014,Dube2017,Dube2020,Shan2021} and global-feature based methods~\cite{He2016,Kim2018,Wang2020}.
Local-feature-based methods either extract geometric and intensity features from local regions or extract high-level object segments from scenes.
However, because of LiDAR's inadequacy to perceive details in local areas, local-feature-based methods undergo low repeatability of extracted features, resulting in limited performance.
In contrast, global-feature-based methods often exploit projections of the point cloud and create signatures for the whole scene, which stress more on the structural information and hence are more effective.
However, existing global descriptors are viewpoint-dependent.
Since LiDARs are usually mounted vertically on top of vehicles, the vehicles drive on roads that can be seen as moving on the planes.
It means that viewpoint changes mainly consist of translation on the $x$-$y$ plane and rotation around the $z$ axis.
Based on these observations, the egocentric polar coordinate system is exploited to overcome the rotation dependence~\cite{Kim2018,Wang2020}.
However, little progress has been made in enabling the translation invariance of global descriptors.

\begin{figure}
\includegraphics[width=1\columnwidth]{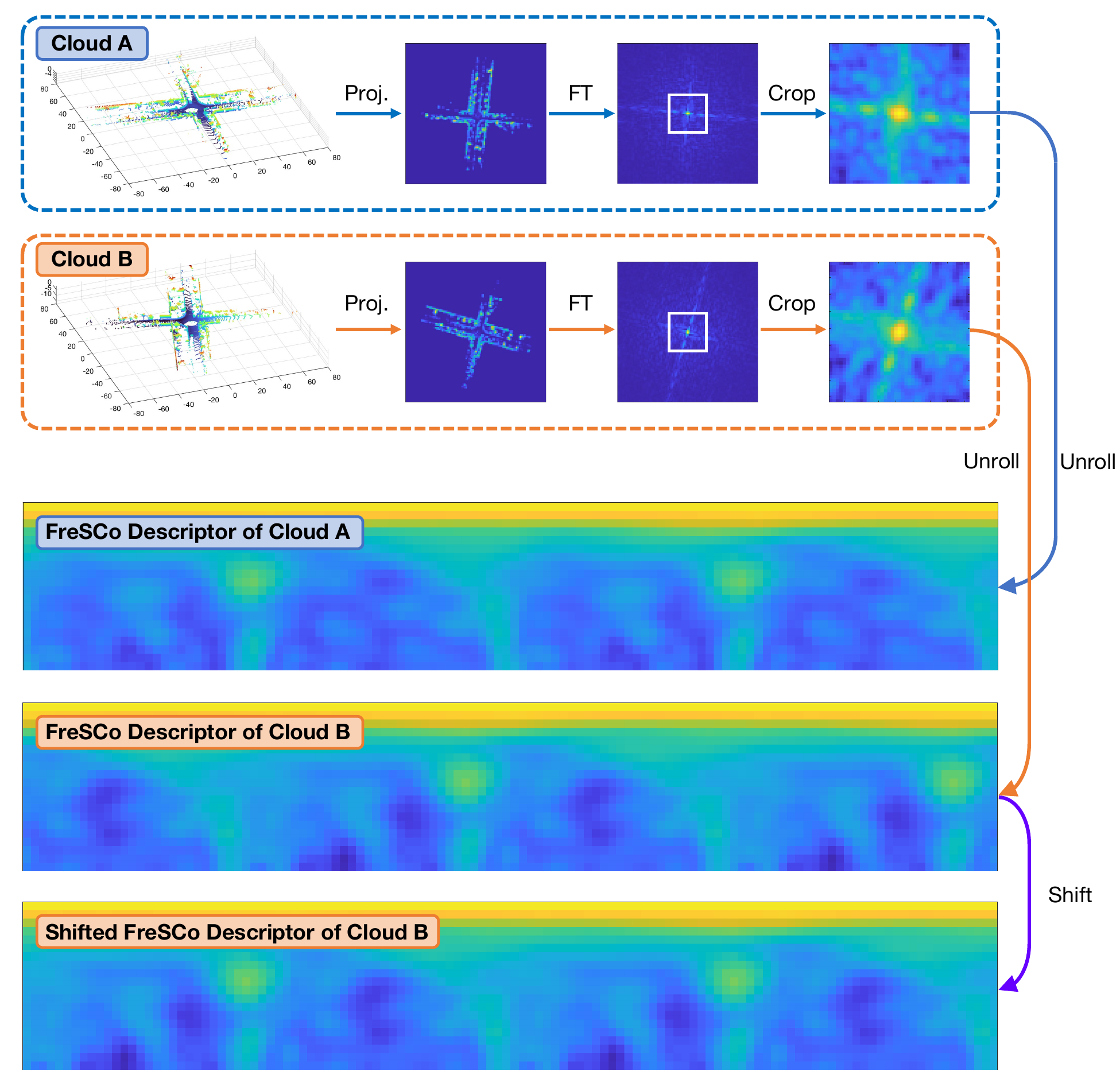}
\caption{
    Illustration of the proposed method performing matching between a query and candidate frame in the place retrieval phase. 
    Here Cloud A and Cloud B are collected around the same place but at different viewpoints.
}
\label{fig_demonstration}
\end{figure}

We present a novel global descriptor called Frequency-Domain Scan Context (FreSCo) for robust place recognition using LiDAR point clouds, which can achieve translation and rotation invariance simultaneously.
As the method name states, the proposed descriptor is introduced in the frequency domain. 
An illustration of the matching mechanism of our method is shown in Fig.~\ref{fig_demonstration}.
Specifically, FreSCo decomposes the relative viewpoint difference between two frames through Fourier Transform. 
Relative translation is eliminated during performing the transform, 
and relative rotation is eliminated by a circular shift operation on the descriptor.
Our proposed FreSCo shows the robust capability of resisting viewpoint changes and captures fine details of the structural information. 
Besides, we also propose a fast two-stage pose estimation method to make FreSCo more practical.
Unlike the commonly used 3D Iterative Closest Point (ICP)-based methods, our proposed method is performed on the compact 2D point cloud, which dramatically reduces the computational cost.

Overall, the main contributions of this paper can be summarized as follows: 
\begin{itemize}
\item A novel global descriptor is introduced in the frequency domain for place retrieval. It can achieve both translation invariance and rotation invariance. 
\item We propose a fast two-stage pose estimation method by leveraging compact 2D point cloud extracted from the original 3D point cloud. 
\item Through evaluation across various datasets, our proposed method surpasses the previous state-of-the-art methods.
\end{itemize}

\section{Related Work}

According to the types of features being exploited, current works on LiDAR-based place recognition can be divided into two categories, that is, local-feature-based methods and global-feature-based methods.

\subsection{Local-Feature-based Methods}

Some approaches exploit the success of local features on 2D images~\cite{Low2004,Bay2006}, and design similar ones adapted for 3D point clouds.
Fast Point Feature Histograms (FPFH)~\cite{Rusu2009} and Signature
of Histograms of Orientations (SHOT)~\cite{Salti2014} build local reference frames at feature points and create signatures of the histogram for the neighboring points.
Intensity Signature of Histograms of Orientations (ISHOT)~\cite{Guo2019} enhances the SHOT descriptor by coupling LiDAR intensity measurements.
These methods rely on point clouds of high density and do not perform well in urban road scenes.
Other works, such as~\cite{Shan2021} and~\cite{DiGiammarino2021}, directly expand visual features to the cylindrical projection of LiDAR intensity readings.
These methods show good invariance to sensor pose changes but need dense-beam LiDAR sensors with high costs.

\subsection{Global-Feature-based Methods}

Global-feature-based methods have shown superior performance on road scenes since they are usually faster and more effective for structural information.
Approaches like Scan Context (SC)~\cite{Kim2018} and its successors, such as Intensity Scan Context (ISC)~\cite{WangHan2020}, LiDAR Iris~\cite{Wang2020} and Differentiable Scan Context with Orientation (DiSCO)~\cite{Xu2021}, utilize projection of point clouds in egocentric polar coordinate systems and generate global descriptor from it.
To tackle the rotation invariance, SC and ISC exploit circular shift technique, while LiDAR Iris and DiSCO leverage Fourier Transform.
However, the ability to resist translation changes is still an open question for these methods.
Seed~\cite{Fan2020}, also a successor to SC, computes a viewpoint-invariant local reference frame on the segmented point cloud to achieve both translation and rotation invariance.
However, the Principle Component Analysis (PCA) operation used during computing the main direction is not robust.
Including DiSCO, many learning-based methods~\cite{Uy2018,Zhang2019,Liu2019,Du2020,zhou2021ndt} also appeared in recent years, and global descriptors are generated using deep neural networks.
Nevertheless, these models have high computational costs, need tremendous data to train, and are prone to degenerate on unseen scenes.

In this paper, we focus on the lightweight approaches that can cover more general computing devices.
The most similar works to ours are SC and LiDAR Iris, and our FreSCo differs in:
1) Our FreSCo utilizes projection of point cloud in the Cartesian coordinate system, free of distortion caused by relative translation, while the polar coordinate system used in theirs is not.
2) In FreSCo, Fourier Transform is exploited to decompose the viewpoint difference between two point clouds and achieve both translation and rotation invariance, while theirs are only rotation-invariant.
3) Our method comes with fast pose estimation, while theirs do not give relative pose unless the computationally heavy 3D ICP is used.

\section{FreSCo}

\begin{figure*}[!t]
    \centering{}
    \includegraphics[width=2.0\columnwidth]{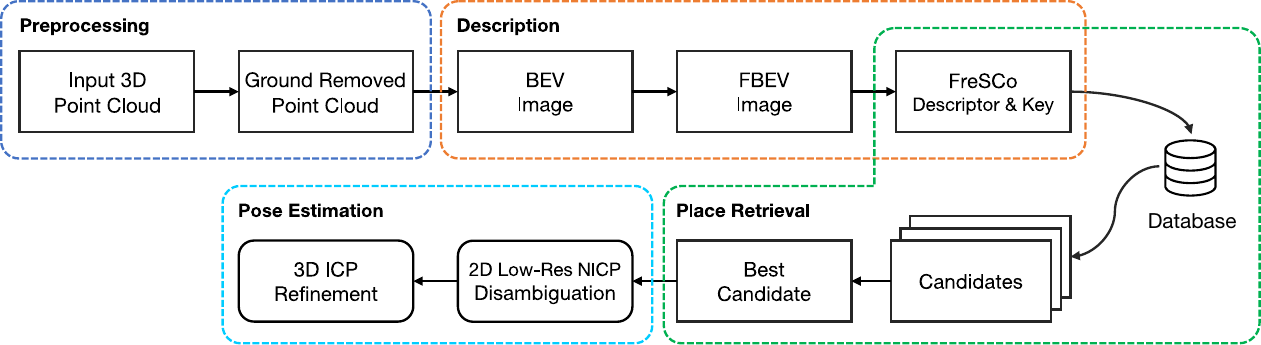}
    \caption{
        Pipeline of the proposed algorithm.
    }
    \label{fig_pipeline}
\end{figure*}

In this section, the proposed method is introduced in detail, and its pipeline is shown in Fig.~\ref{fig_pipeline}.
First, the LiDAR point cloud is projected to achieve the bird's eye view (BEV) image.
The frequency-domain representation of the BEV image is shown to be less sensitive to LiDAR's viewpoint changes.
Then, the frequency-domain representation is utilized to create FreSCo as the translation-invariant and rotation-invariant global descriptor.
Finally, a fast pose estimation method between the query and match keyframes is introduced to obtain the final relative pose.

\subsection{BEV Image Generation}

The idea of leveraging the BEV image is inspired by SC~\cite{Kim2018}, which projects the point cloud onto the 2D plane and encode the maximum height information around the viewpoint in a discrete polar coordinate system.
This encoding manner from SC reserves the robust feature of the surrounding scenes, as the height of buildings does not vary, and its measuring is highly repeatable.
However, its polar representation hinders the invariance to viewpoint changes.
When revisiting the same place, if the vehicle shifts from former viewpoint straightly or sideways, the projected image can distort significantly, making it hard to recognize the revisit.
To this end, we encode the height information in the Cartesian coordinate system, ensuring the projected image does not get distorted when translation occurs during revisit.

In short, we bin the point cloud based on their $x$-$y$ coordinates and then count the $z$-axis maxima of each bin.
Let $P=\{p_{1},p_{2},...,p_{n}\}$ be the point cloud acquired at a keyframe, where $p_{i}=[x_{i},y_{i,}z_{i}]^{T}$ is a 3D point in $P$. 
Suppose that the LiDAR is mounted vertically on top of the vehicle, with its $z$ axis pointing upward. 
We consider a squared-shape window with $S\times S$ bins, covering an area of size $L\times L$
in the $x$-$y$ plane, and centered at the origin of the point cloud.
We assign all the points within the covered area to their belonging bins. For point $i$, its belonging bin is given by:
\begin{equation}
r_{i}=\ensuremath{[\frac{x_{i}+L/2}{L/S}]},c_{i}=[\frac{y_{i}+L/2}{L/S}],
\end{equation}
where $[\cdot]$ is the rounding operation, and $r_{i}$ and $c_{i}$ are the row index and col index of the belonging bin. 
And then BEV image is generated using the maximum $z$ value of points within each bin. 
Considering that ground points in the point cloud cannot provide distinctive information in the place recognition task, and that it causes unwanted intensity changes in the BEV image even when the vehicle tilts with a small pitch or roll angle, we remove all ground points using method~\cite{Himmelsbach2010} before generating the BEV image. 

\subsection{Frequency-Domain Representation}

Although we have obtained the BEV image which is free of distortion caused by viewpoint changes, the translation and rotation changes still remain. 
To achieve a truly viewpoint-invariant global descriptor, we adopt 2D Fourier Transform of the BEV (FBEV) image to eliminate such changes. 

Given two BEV images $I_{1}^{B}$ and $I_{2}^{B}$ only differ by a displacement $(\delta_{x},\delta_{y})$ caused by the
translation during revisit, \emph{i.e.}, $I_{2}^{B}(x,y)=I_{1}^{B}(x-\delta_{x},y-\delta_{y})$.
Denote their FBEV images as $I_{1}^{F}$ and $I_{2}^{F}$, we have 
\begin{equation}
I_{2}^{F}(u,v)=e^{i2\pi(\delta_{x}u+\delta_{y}v)}\cdot I_{1}^{F}(u,v).
\end{equation}
Since the norm of $e^{i2\pi(\delta_{x}u+\delta_{y}v)}$ is equal to 1, we can derive
\begin{equation}
\|I_{2}^{F}(u,v)\|=\|I_{1}^{F}(u,v)\|,
\end{equation}
where $\| \cdot \|$ is the magnitude of the FBEV image.
It indicates that a translation in the $x$-$y$ plane of the point cloud no longer makes a difference to the FBEV image. 

Suppose $I_{1}^{B}$ and $I_{2}^{B}$ only differ by a rotation $\delta_{\theta}$ caused by the vehicle rotation during revisit.
In polar coordinates, the spacial-domain and frequency-domain coordinates are: 
\begin{equation}
\begin{cases}
x=r{\rm cos}\theta\\
y=r{\rm sin}\theta
\end{cases}
\begin{cases}
u=\rho{\rm cos}\phi\\
v=\rho{\rm sin\phi}
\end{cases},
\end{equation}
such that $I_{2}^{B}(r,\theta)=I_{1}^{B}(r,\theta+\delta_{\theta})$.
Then we have 
\begin{equation}
I_{2}^{F}(\rho,\phi)=I_{1}^{F}(\rho,\phi+\delta_{\theta}).
\end{equation}
Still, we take the norm on both sides, which leads to:
\begin{equation}
\text{\ensuremath{\|}}I_{2}^{F}(\rho,\phi)\|=\text{\ensuremath{\|}}I_{1}^{F}(\rho,\phi+\delta_{\theta})\|.
\end{equation}
It indicates that a rotation in the yaw axis in the spacial-domain results in the same rotation in the frequency domain. 

\begin{figure}
\begin{centering}
\subfloat[]{
\includegraphics[width=0.22\columnwidth]{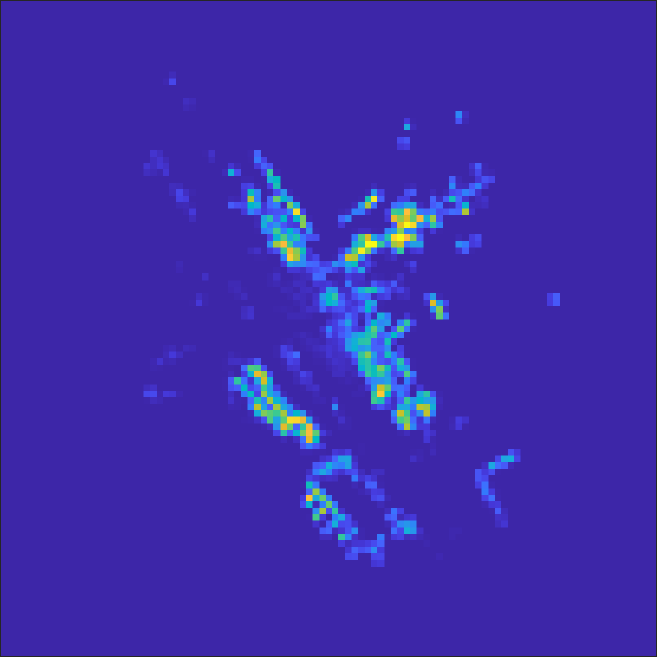}
}
\subfloat[]{
\includegraphics[width=0.22\columnwidth]{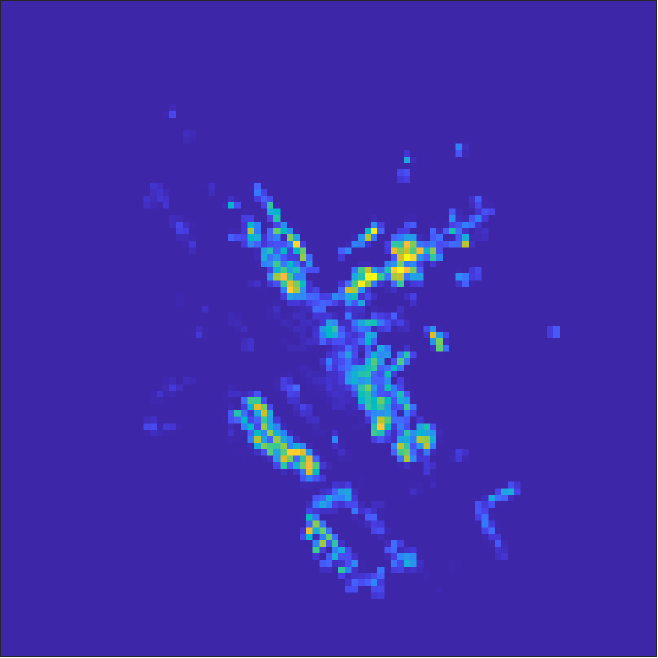}
}
\subfloat[]{
\includegraphics[width=0.22\columnwidth]{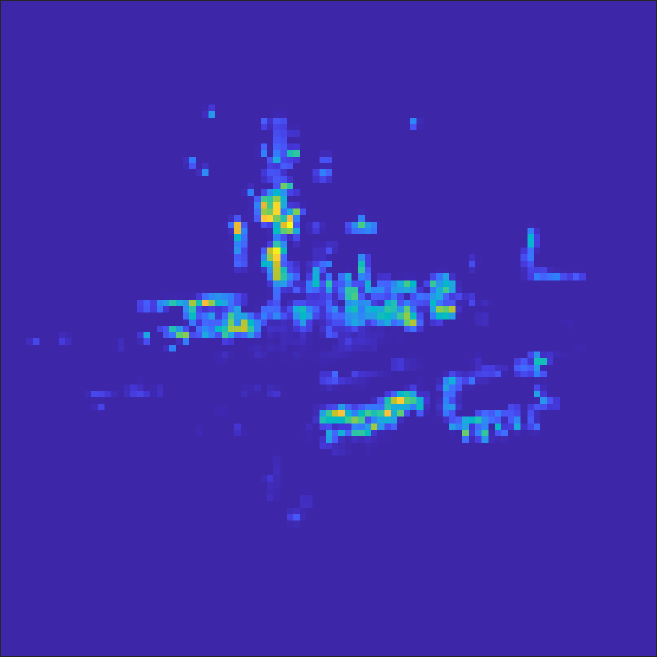}
}
\subfloat[]{
\includegraphics[width=0.22\columnwidth]{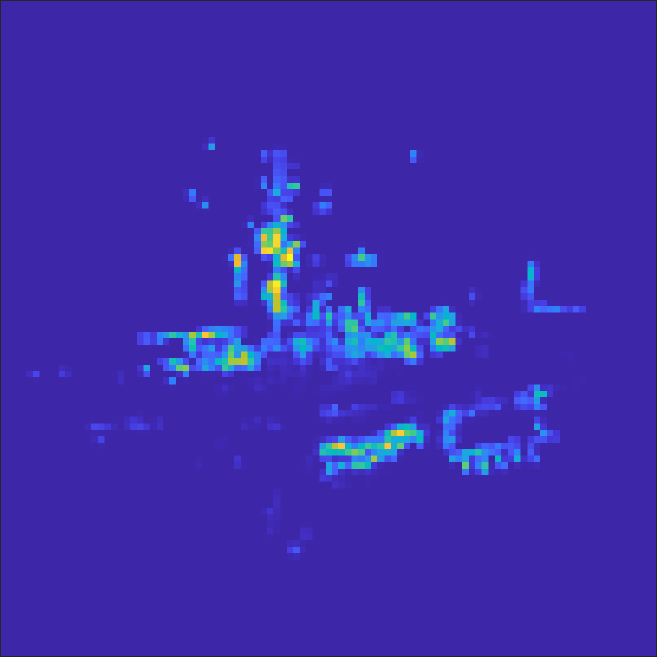}
}
\par\end{centering}
\centering{}
\subfloat[]{
\includegraphics[width=0.22\columnwidth]{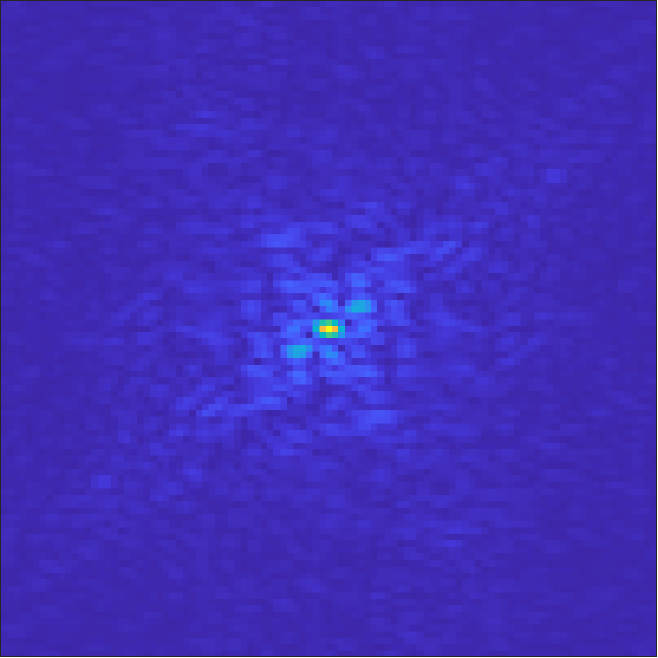}
}
\subfloat[]{
\includegraphics[width=0.22\columnwidth]{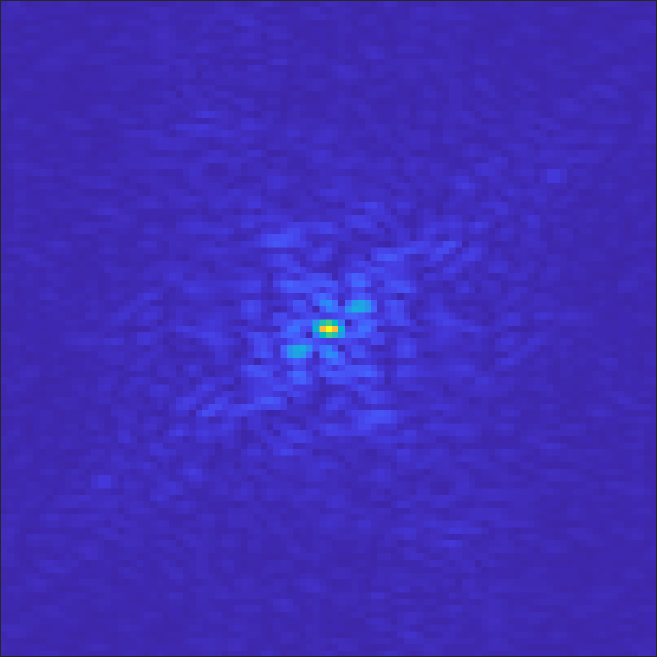}
}
\subfloat[]{
\includegraphics[width=0.22\columnwidth]{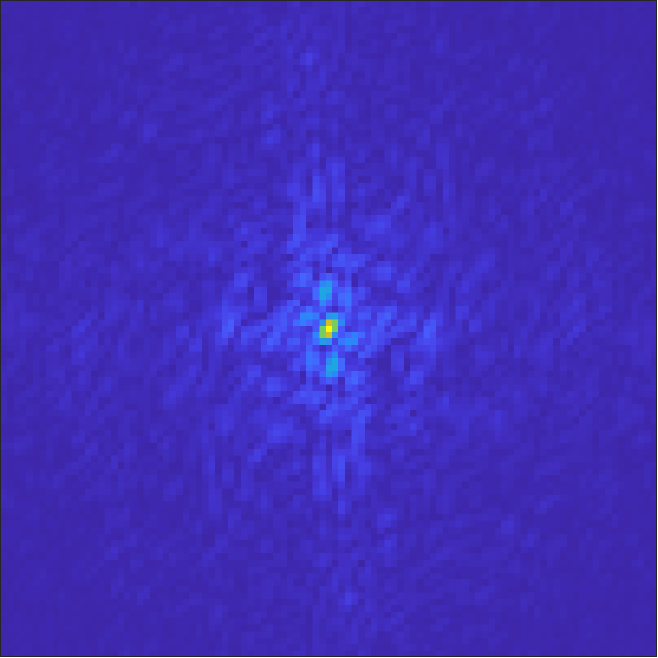}
}
\subfloat[]{
\includegraphics[width=0.22\columnwidth]{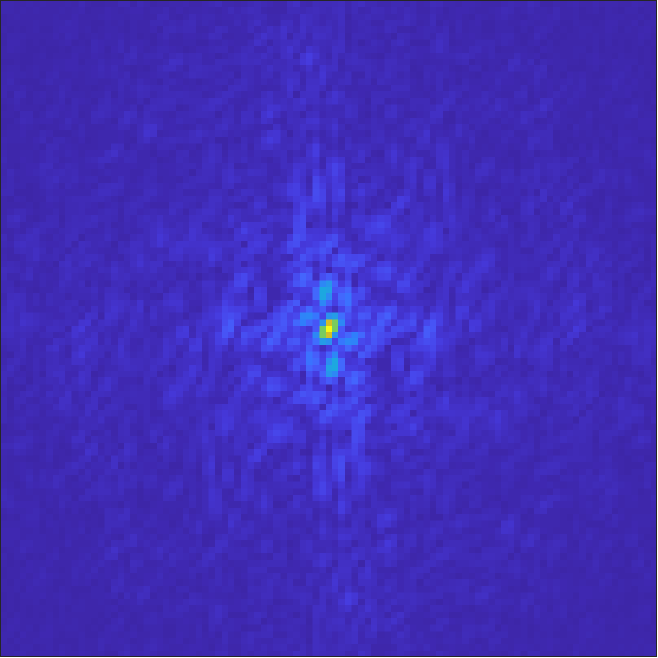}
}
\caption{
    The BEV images (first row) and their corresponding FBEV images (second row) under different viewpoint changes. 
    From left to right are the results on the original point cloud, shifted 10m on the $x$ axis, rotated $60{^\circ}$ around the $z$ axis, and the combination of the above two, respectively.
}
\label{fig_ft_property}
\end{figure}

So far, by presenting the BEV image in the frequency domain, the difference in viewpoint is decomposed into first relative translation and then relative rotation. 
The translation is eliminated using the Fourier Transform, and the rotation is kept equivariant after the transformation.
Such properties are exhibited in Fig.~\ref{fig_ft_property}.

\subsection{FreSCo Descriptor Generation and Matching}

We note that low-frequency components are the most distinguishable in the FBEV image, so we crop the center part and use it as the signature for the surrounding scenes of the keyframe. 
Beside, we take the logarithm of all values in the cropped area to prevent the dc component from dominating the signature. 

FreSCo descriptor is a two-dimensional array of size $N_{v}\times N_{h}$ obtained by unrolling the cropped FBEV image in the polar coordinate system. 
Its vertical and horizontal axes correspond to the radial and angular distance in the frequency domain, respectively. 
Since the rotation on the FBEV image is consistent with that of the point cloud, a rotation on the point cloud can be
interpreted as a circular shift on the FreSCo descriptor's horizontal axis. 

For a query FreSCo descriptor $\mathscr{D}_{q}$ and a candidate FreSCo descriptor $\mathscr{D}_{c}$, we use two distance measurements to characterize the similarity between them. 
First, $\mathscr{D}_{c}$ is circularly shifted in the horizontal axis to achieve its best shift $\delta_{h}^{*}$, such that the $L_{1}$ distance between $\mathscr{D}_{q}$ and shifted $\mathscr{D}_{q}$ reaches the minimum, which is formulated as: 
\begin{equation}
\delta_{h}^{*}=\mathop{\arg\min}_{\delta_{h}\in[0,\frac{N_{h}}{2}-1]}\|\mathscr{D}_{q}-f(\mathscr{D}_{c},\delta_{h})\text{\ensuremath{\|}}_{1},
\end{equation}
where $f(\mathscr{D}_{c},\delta_{h})$ is the operator that circularly shifts $\mathscr{D}_{c}$ in the horizontal axis by $\delta_{h}$.
Since the FBEV image is symmetric with respect to its center, the FreSCo descriptor repeats itself horizontally. 
In other words, circular shift on $\mathscr{D}_{c}$ is only needed within half of its maximum range. 
The first distance measurement $d_{L_{1}}$ is given by: 
\begin{equation}
d_{L_{1}}=\|\mathscr{D}_{q}-f(\mathscr{D}_{c},\delta_{h}^{*})\text{\ensuremath{\|}}_{1}.
\end{equation}
This measurement describes the best overall distance between $\mathscr{D}_{q}$ and $\mathscr{D}_{c}$. 

Next, the row-wise cosine distance $d_{R}$ between $\mathscr{D}_{q}$ and $\mathscr{D}_{c}$ is computed, which is given by: 
\begin{equation}
d_{R}=1-\frac{1}{N_{v}}\sum_{i=1}^{N_{v}}\frac{r_{i}^{q}\cdot r_{i}^{c}}{\|r_{i}^{q}\|\|r_{i}^{c}\|},
\end{equation}
where $r_{i}^{q}$ and $r_{i}^{c}$ represents the $i^{{\rm th}}$ row of $\mathscr{D}_{q}$ and shifted $\mathscr{D}_{c}$, respectively.
$d_{R}$ describes the difference of orientations on different levels of frequency components between two FreSCo descriptors. 

\subsection{Fast Matching}

To ensure fast matching for each query, a key is generated for each FreSCo descriptor and utilized to select candidate descriptors.
The key is a vector of size $2N_{v}$, and is computed on each row $r$ of the FreSCo descriptor $\mathscr{D}$ by:
\begin{equation}
\mathscr{K}=\frac{1}{{\mu}(\mathscr{D})}[{\mu}(r_{1}),...,{\mu}(r_{N_{v}}), {\sigma}(r_{1}),...,{\sigma}(r_{N_{v}})]^{T},
\end{equation}
where ${\mu}(\cdot)$ represents the mean value of the matrix or vector, ${\sigma(\cdot)}$ represents the standard deviation of the vector. 
The key is normalized over ${\mu}(\mathscr{D})$ to enhance its robustness to noise. 

All keys are stored in the k-D tree for future nearest neighbors search.
For a query FreSCo, its key is extracted and used to find nearest $N_{c}$ neighbors in the k-D tree, and the corresponding FreSCo descriptors of these neighbors compose the match candidates for the query FreSCo.
The $d_{L_{1}}$ distance is calculated for each candidate, and the one with the smallest $d_{L_{1}}$ distance is selected as the best candidate, which is finally verified on the orientation of different frequency components using the $d_{R}$ distance.

\begin{figure}[h]
	\centering{}
	\subfloat[]{\includegraphics[width=0.32\columnwidth]{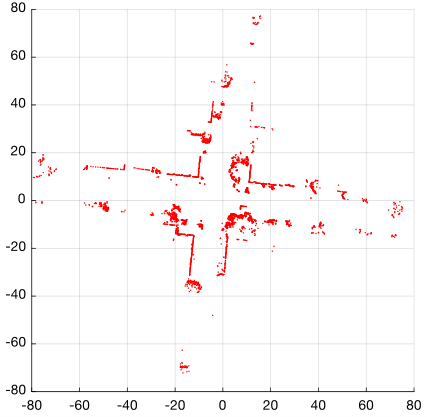}}
	\subfloat[]{\includegraphics[width=0.32\columnwidth]{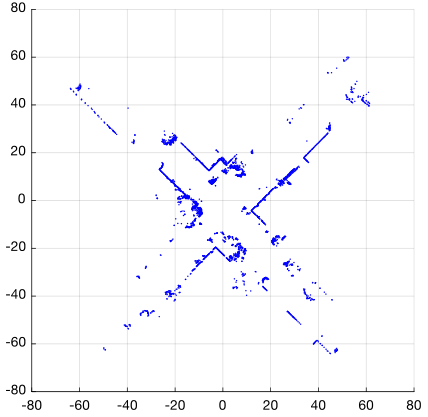}}
	\subfloat[]{\includegraphics[width=0.32\columnwidth]{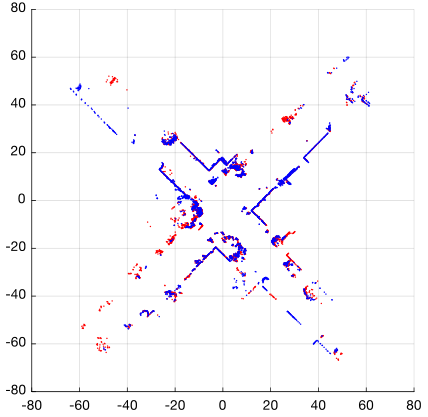}}
	\caption{An example of the first stage pose estimation.
		(a) and (b) are top views of the compact 2D point clouds extracted from
		$1512^{{\rm th}}$ and $714^{{\rm th}}$ frame from KITTI \texttt{08},
		respectively. (c) corresponds to the result of (a) aligned to (b)
		using the estimated relative pose.}
	\label{fig_pose_estimation}
\end{figure}

\subsection{Fast Pose Estimation}

The aim of place retrieval using the global descriptor is to find
the coarse location in the global frame. Point cloud registration
methods should be used to estimate the relative pose in a finer scale.
3D ICP is often exploited to align two point
clouds and estimate the relative pose, while it is computationally
heavy and sensitive to initial values. To tackle such problem, a two-stage
fast pose estimation procedure is proposed. 

Note that when verifying the candidate in the matching phase, a best
shift $\delta_{h}^{*}$ on the FreSCo descriptor is achieved, corresponding
to a rotation $\delta_{\theta}^{*}$ in the FBEV image: 

\begin{equation}
	\delta_{\theta}^{*}=\frac{\delta_{h}^{*}}{N_{h}}\times360{^\circ}
\end{equation}
Recall that a rotation in the point cloud leads to the same rotation
on the FBEV image. However, due to the centro-symmetric nature of
the FBEV image, the rotation $\delta_{\theta}^{*}$ may result from
two circumstances of different rotations in the point cloud, which
leads to an ambiguity when estimating the relative rotation. To eliminate
such ambiguity, 2D ICP with Normals (NICP) is exploited. Since planar
and curved surfaces are commonly perpendicular to the ground in urban
road scenes, the structure information can be well reserved when the
point clouds are flattened to the $x$-$y$ plane. Therefore, 2D NICP
can be effective for the first-stage pose estimation. First, for the
ground-removed 3D point cloud, its higher-part points are extracted and then flattened on the $x$-$y$ plane to achieve a compact 2D point
cloud. To make sure the points are uniformly extracted, the $x$-$y$
plane is divided into coarse grids beforehand, and the extraction
is performed on each grid individually. Then, the 2D point cloud is
down-sampled to a low resolution to ensure the speed. Next, we run
2D NICP on the 2D point clouds from the query scan and match scan
twice, with initial guesses for the rotation angle $\delta_{\theta}^{*}$
and $\delta_{\theta}^{*}+180{^\circ}$, respectively. The result with
a smaller mean square error (MSE) is considered to be the proper relative
pose. An example of the first stage pose estimation is shown in Figure
\ref{fig_pose_estimation}. 

The precision of the relative pose can be optionally refined by running 3D ICP on the 3D point clouds with formerly obtained relative pose as initial guess.

\section{Experimental Evaluation}

\begin{figure*}[t]
\centering{}
\subfloat[MulRan \texttt{KAIST 03}]{\includegraphics[width=0.27\paperwidth]{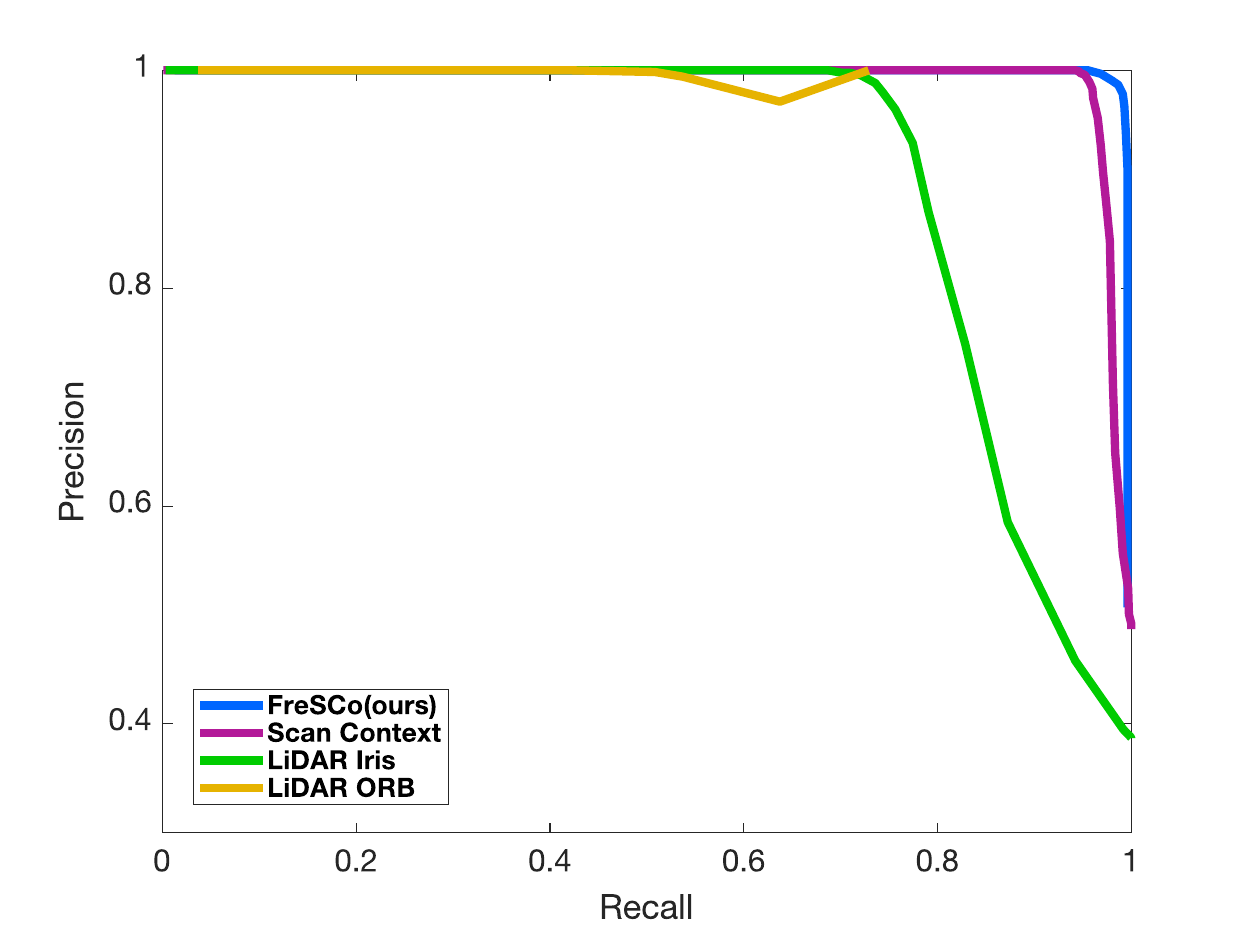}}
\subfloat[KITTI \texttt{08}]{\includegraphics[width=0.27\paperwidth]{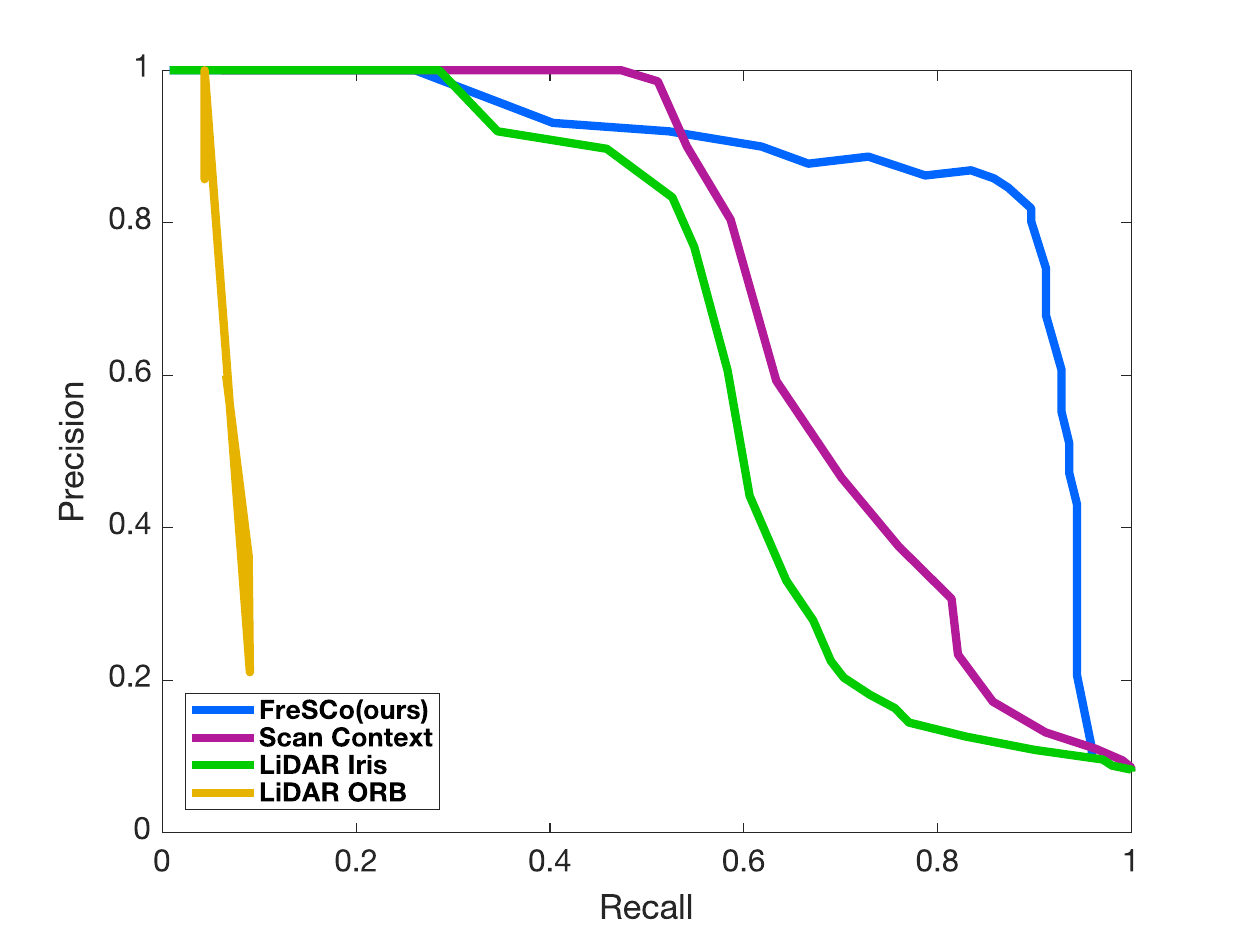}}
\subfloat[Oxford \texttt{2019-01-11-13-24-51}]{\includegraphics[width=0.27\paperwidth]{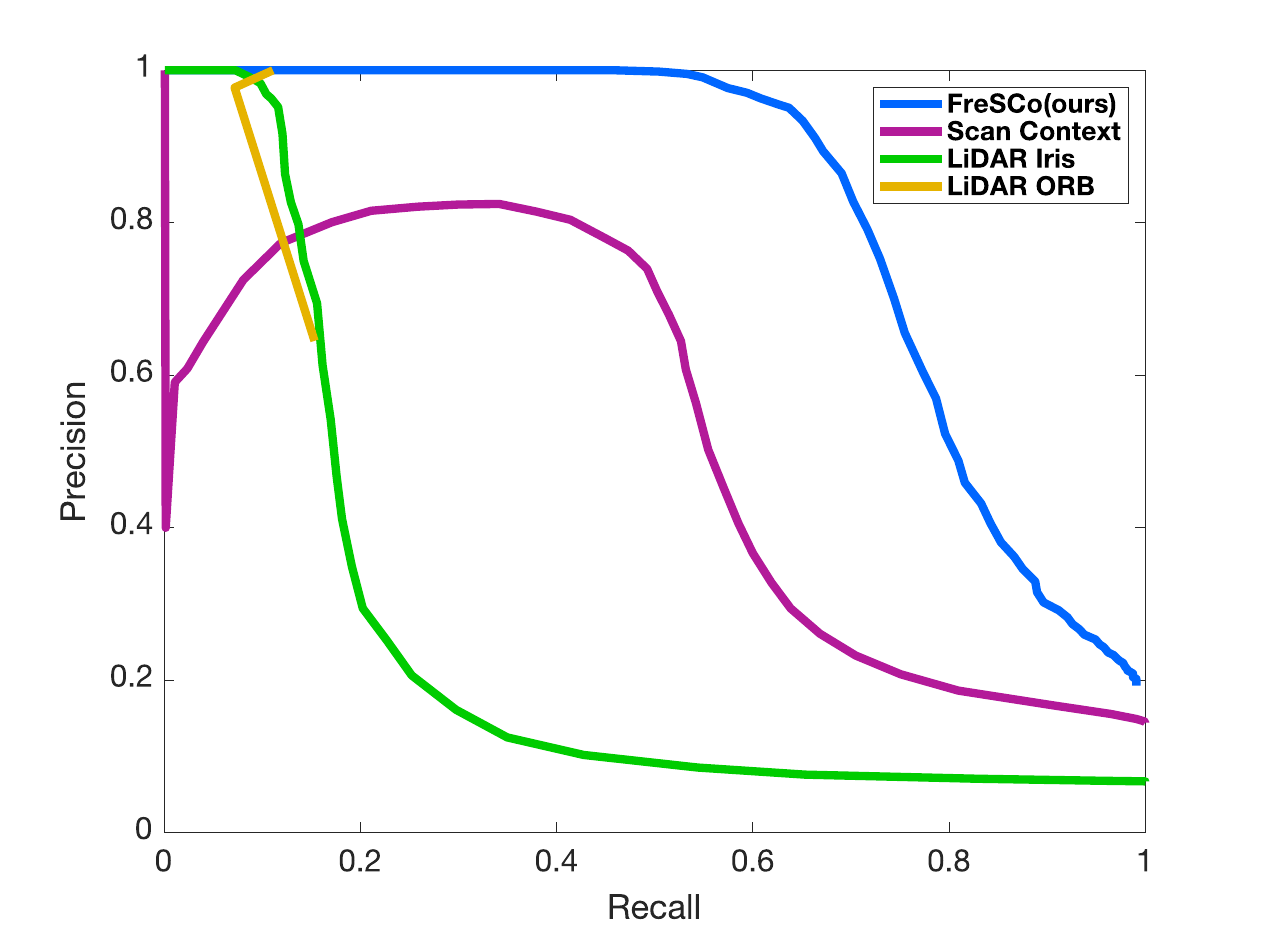}}
\caption{Precision-recall curves on different dataset sequences.}
\label{fig_pr_curves}
\end{figure*}

In this section, the performance of the proposed method is evaluated.
First, we compared the place retrieval performance of FreSCo with some state-of-the-art methods, that is, Scan Context~\cite{Kim2018}, LiDAR Iris~\cite{Wang2020} and LiDAR ORB~\cite{Shan2021}, among which the first two methods are based on global descriptor, and the last one is based on local descriptor. 
Then, we evaluated the accuracy and success rate of the proposed fast pose estimation method compared to 3D ICP. 
Finally, we showed the comparison of overall runtime. 

\subsection{Datasets and Experimental Setups}

Experiments are conducted on three datasets of various scenes and various sensor types: MulRan dataset~\cite{KimMulRan2020}, KITTI dataset~\cite{GeigerKitti2012} and Oxford Radar RobotCar dataset~\cite{BarnesOxfordRadar2020}. 

\textit{1) MulRan dataset}: 
The \texttt{DCC} and \texttt{KAIST} sequences of MulRan dataset provide measurements collected inside research institutes using a 64-beam LiDAR Ouster OS1-64. 
This dataset has a high diversity of structures. 
Considering that LiDAR point clouds in this dataset always have a blind region of about $71{^\circ}$ caused by the radar sensor mounted behind it, the algorithm's performance on sequences containing reversed loops can be severely impacted. 
Therefore, sequence \texttt{KAIST 03}, which only contains loops of the same direction, is selected. 

\textit{2) KITTI dataset}: 
The KITTI dataset is collected in a residential area using a 64-beam LiDAR Velodyne HDL-64E. 
The concentration of beams in the downward direction causes an absence of structural details from buildings at higher altitudes, making it challenging for place recognition tasks. 
We select sequence \texttt{08} of the dataset, as this sequence only contains reversed loops and also has a bigger translation during revisit. 

\textit{3) Oxford Radar RobotCar dataset}: 
The Oxford Radar RobotCar dataset is collected by repeatedly traversing a route in Oxford through various weather conditions. 
The point clouds are obtained using two 32-beam LiDARs Velodyne HDL-32E.  
We only use data from the left LiDAR. 
Sequence \texttt{2019-01-11-13-24-51} is selected, as it has fine GPS coverage throughout almost the whole trajectory. 

The source codes of the methods for comparison are acquired through their open-source implementations.
Although the LiDAR Iris is based on global descriptor, unlike Scan Context and our FreSCo, it achieves retrieval through traversing all previous descriptors and performing brute-force matching on each of them, which
is not feasible for real-time evaluation. 
For the sake of fairness, we implemented fast matching for LiDAR Iris by generating Scan-Context-style keys for its descriptors. 
The key is a vector composed of the number of ones in each row of its binary descriptor. 

Tests are run on keyframes sampled from the datasets, and distance is set to $2{\rm m}$ between two consecutive keyframes. 
When evaluated on ground truth, a matching pair of a query keyframe and a previous keyframe within $10{\rm m}$ of it, is considered to be a true positive.
We use 20 candidates for Scan Context, LiDAR Iris, and our proposed method. 
Other parameters of the methods for comparison are kept default. 

\subsection{Place Retrieval}

\begin{table}
\centering{}%
\caption{\label{tab_comparison}Comparison on place retrieval with existing methods.}
\begin{tabular}{ll>{\centering}p{1cm}>{\centering}p{1cm}>{\centering}p{1cm}}
\toprule 
\multirow{2}{*}{Dataset} & \multirow{2}{*}{Approach} & Max F1 & Precision & Recall\tabularnewline
&  & Score & (\%) & (\%)\tabularnewline
\midrule 
& Scan Context & 0.9735 & 99.59 & 95.20\tabularnewline
\cmidrule{2-5} \cmidrule{3-5} \cmidrule{4-5} \cmidrule{5-5} 
MulRan & LiDAR Iris & 0.8478 & 96.42 & 75.64\tabularnewline
\cmidrule{2-5} \cmidrule{3-5} \cmidrule{4-5} \cmidrule{5-5} 
\texttt{KAIST 03} & LiDAR ORB & 0.8432 & 100.00 & 72.89\tabularnewline
\cmidrule{2-5} \cmidrule{3-5} \cmidrule{4-5} \cmidrule{5-5} 
& \textbf{FreSCo (ours)} & \textbf{0.9866} & \textbf{98.66} & \textbf{98.66}\tabularnewline
\midrule 
& Scan Context & 0.6783 & 80.41 & 58.65\tabularnewline
\cmidrule{2-5} \cmidrule{3-5} \cmidrule{4-5} \cmidrule{5-5} 
KITTI & LiDAR Iris & 0.6452 & 83.33 & 52.63\tabularnewline
\cmidrule{2-5} \cmidrule{3-5} \cmidrule{4-5} \cmidrule{5-5} 
\texttt{08} & LiDAR ORB & 0.1429 & 36.36 & 8.89\tabularnewline
\cmidrule{2-5} \cmidrule{3-5} \cmidrule{4-5} \cmidrule{5-5} 
& \textbf{FreSCo (ours)} & \textbf{0.8594} & \textbf{84.62} & \textbf{87.30}\tabularnewline
\midrule 
& Scan Context & 0.5911 & 73.97 & 49.23\tabularnewline
\cmidrule{2-5} \cmidrule{3-5} \cmidrule{4-5} \cmidrule{5-5} 
Oxford & LiDAR Iris & 0.2586 & 54.22 & 16.98\tabularnewline
\cmidrule{2-5} \cmidrule{3-5} \cmidrule{4-5} \cmidrule{5-5} 
\texttt{\footnotesize{}2019-01-11} & LiDAR ORB & 0.2478 & 64.52 & 15.33\tabularnewline
\cmidrule{2-5} \cmidrule{3-5} \cmidrule{4-5} \cmidrule{5-5} 
\texttt{\footnotesize{}-13-24-51} & \textbf{FreSCo (ours)} & \textbf{0.7679} & \textbf{86.49} & \textbf{69.05}\tabularnewline
\bottomrule
\end{tabular}
\end{table}

\begin{figure}[t]
\begin{centering}
\subfloat[FreSCo (ours)]{\includegraphics[width=0.2\paperwidth]{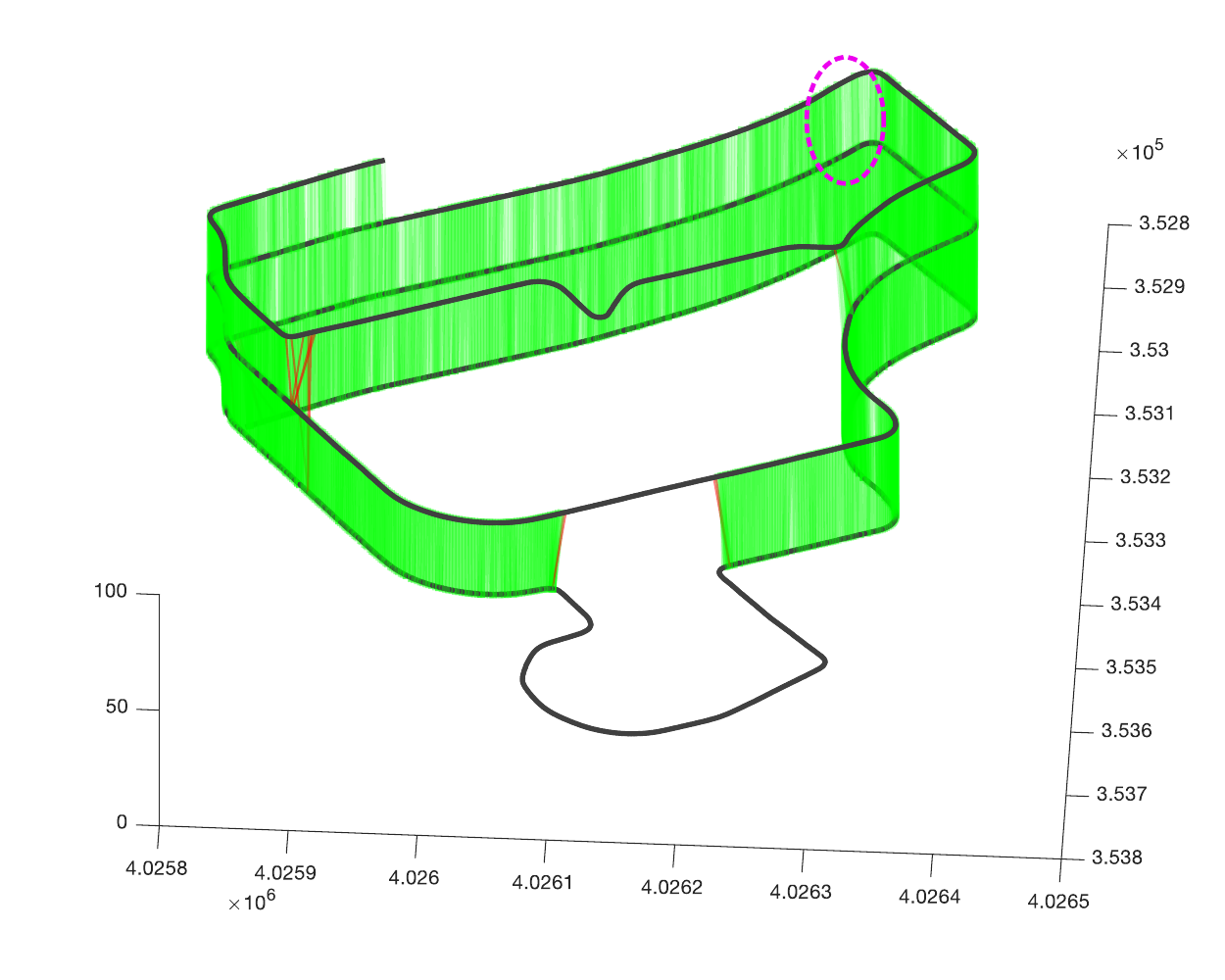}
}\subfloat[Scan Context]{\includegraphics[width=0.2\paperwidth]{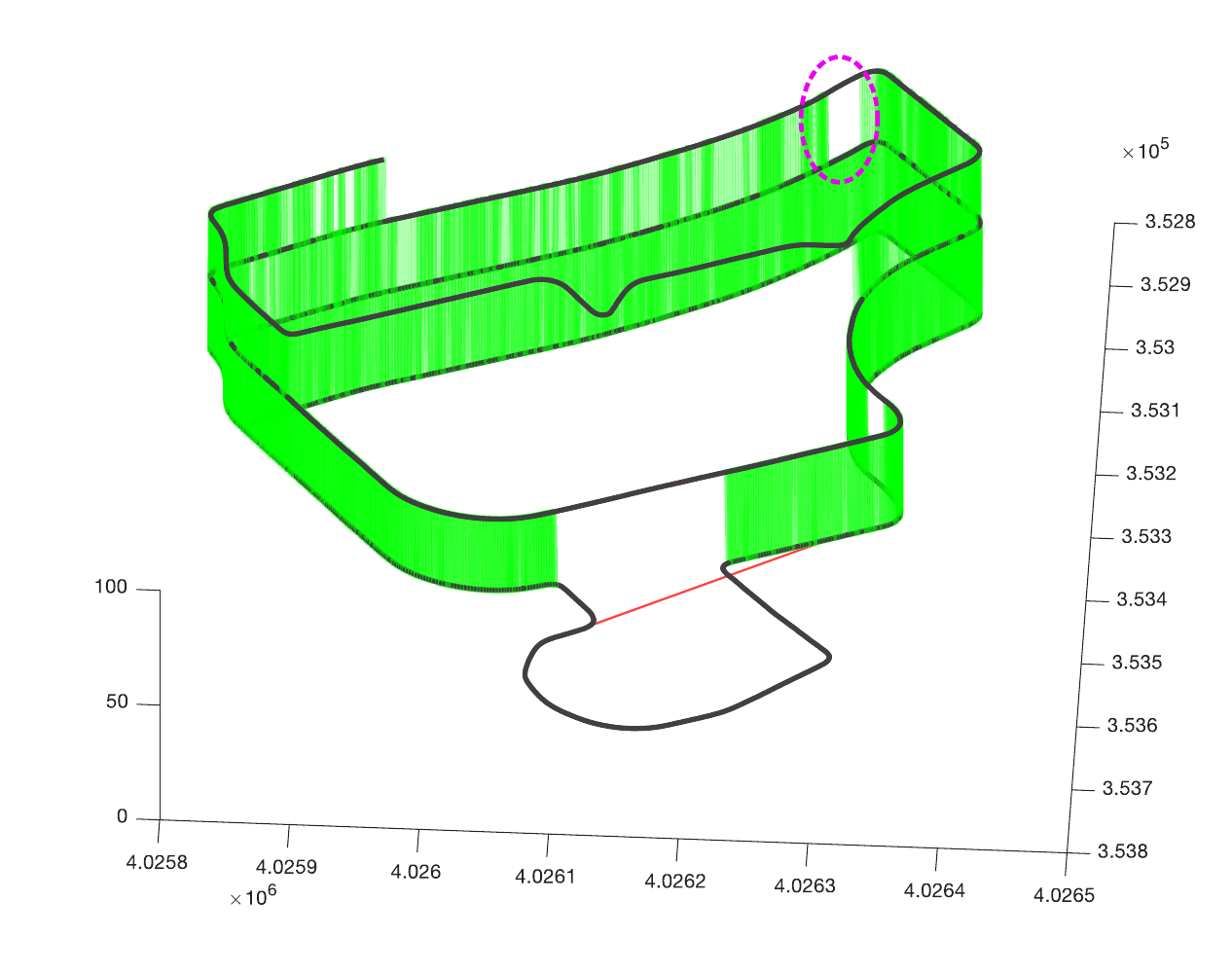}

}
\par\end{centering}
\centering{}
\subfloat[LiDAR Iris]{\includegraphics[width=0.2\paperwidth]{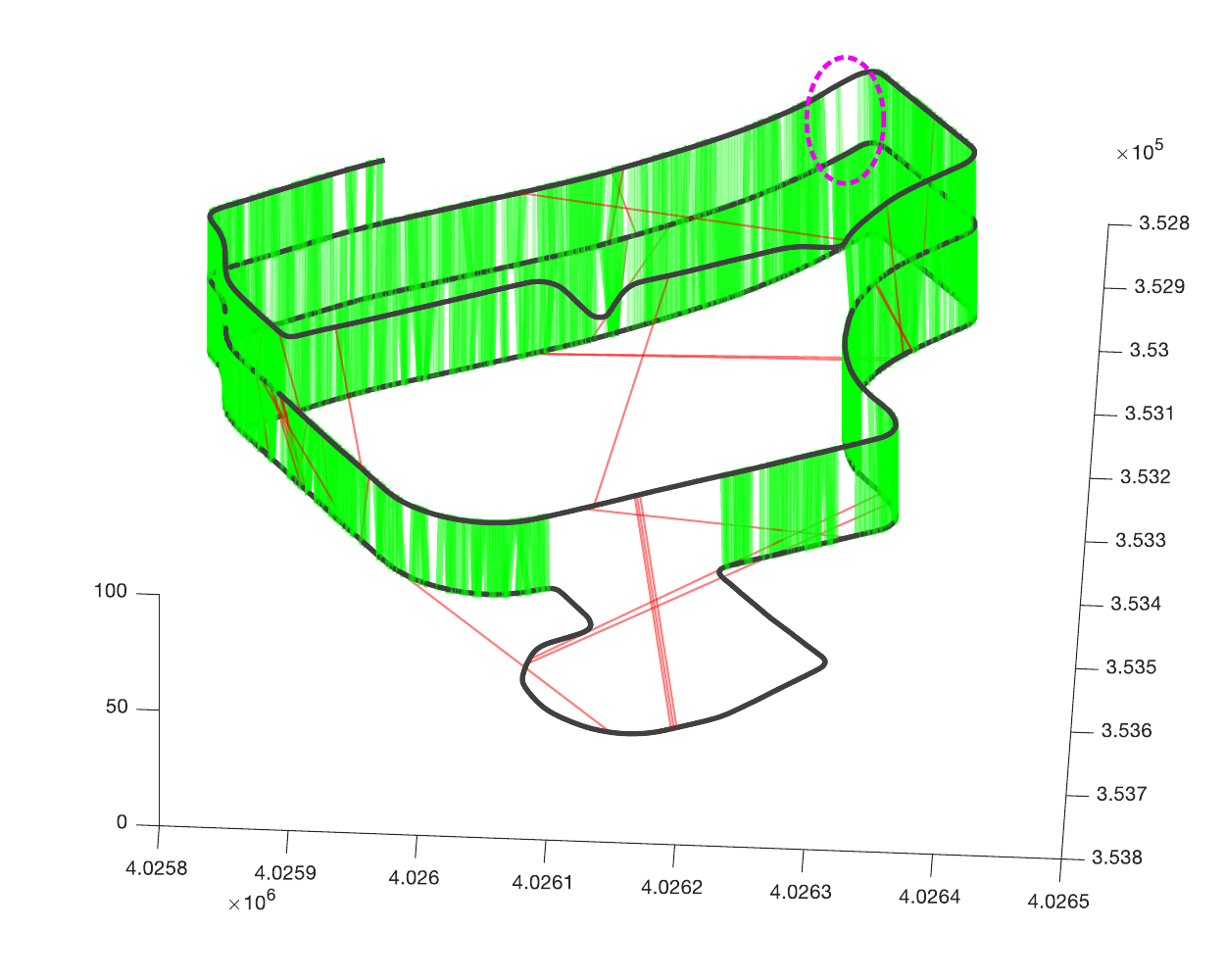}}
\subfloat[LiDAR ORB]{\includegraphics[width=0.2\paperwidth]{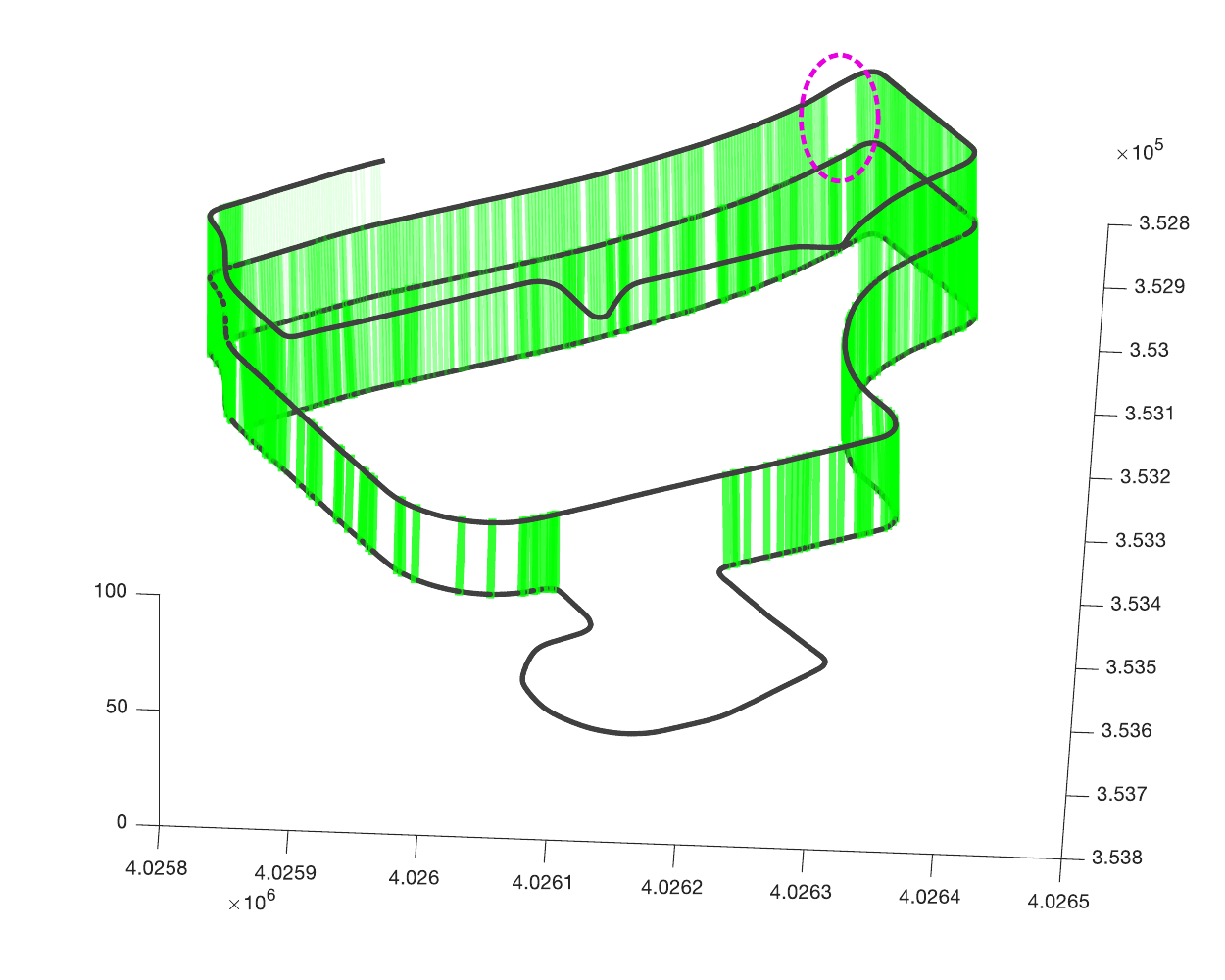}}\caption{\label{fig_result_mulran}Place recognition result on MulRan dataset.}
\end{figure}

\begin{figure}
\begin{centering}
\subfloat[FreSCo (ours)]{\includegraphics[width=0.2\paperwidth]{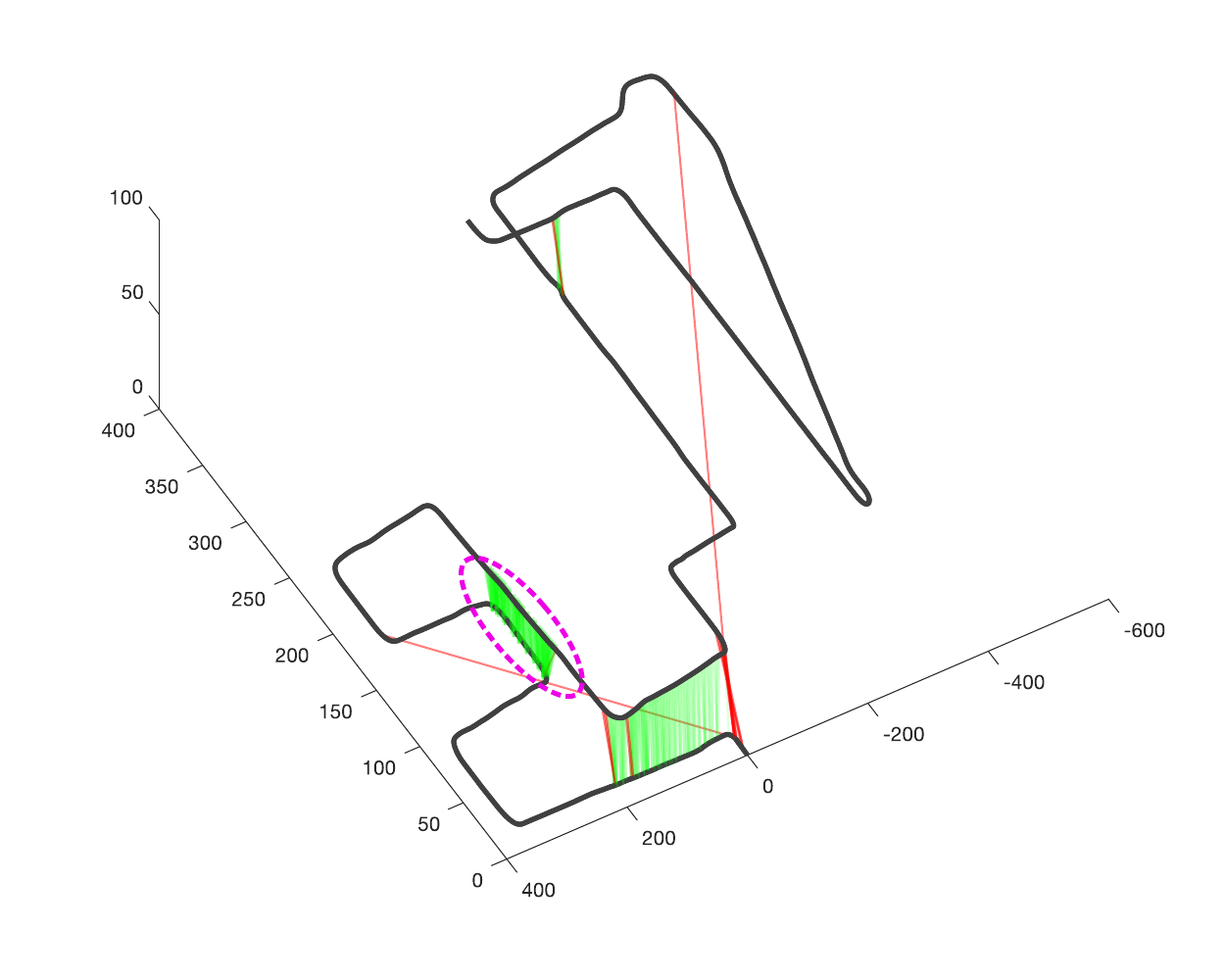}
}
\subfloat[Scan Context]{\includegraphics[width=0.2\paperwidth]{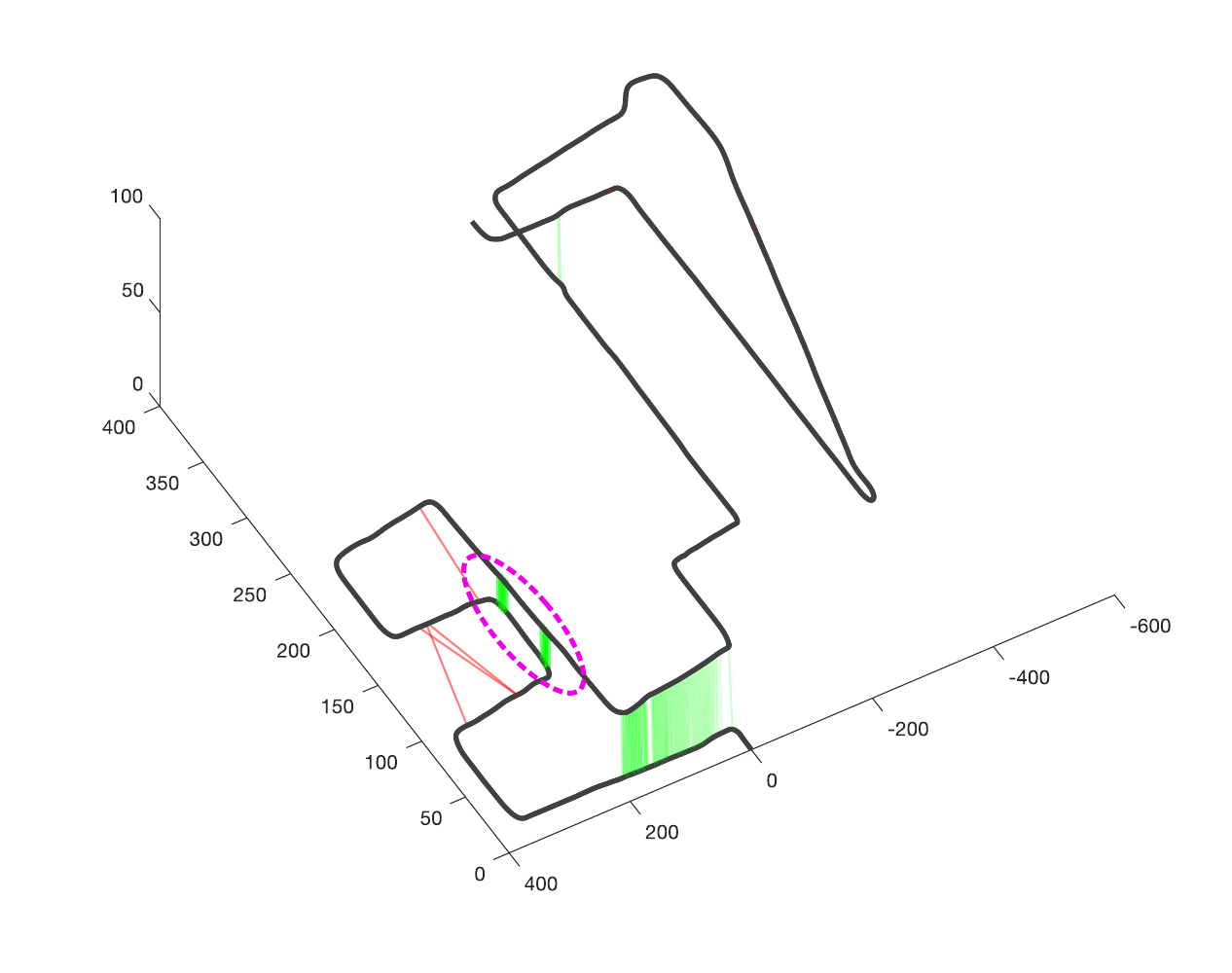}
}
\par\end{centering}
\centering{}
\subfloat[LiDAR Iris]{\includegraphics[width=0.2\paperwidth]{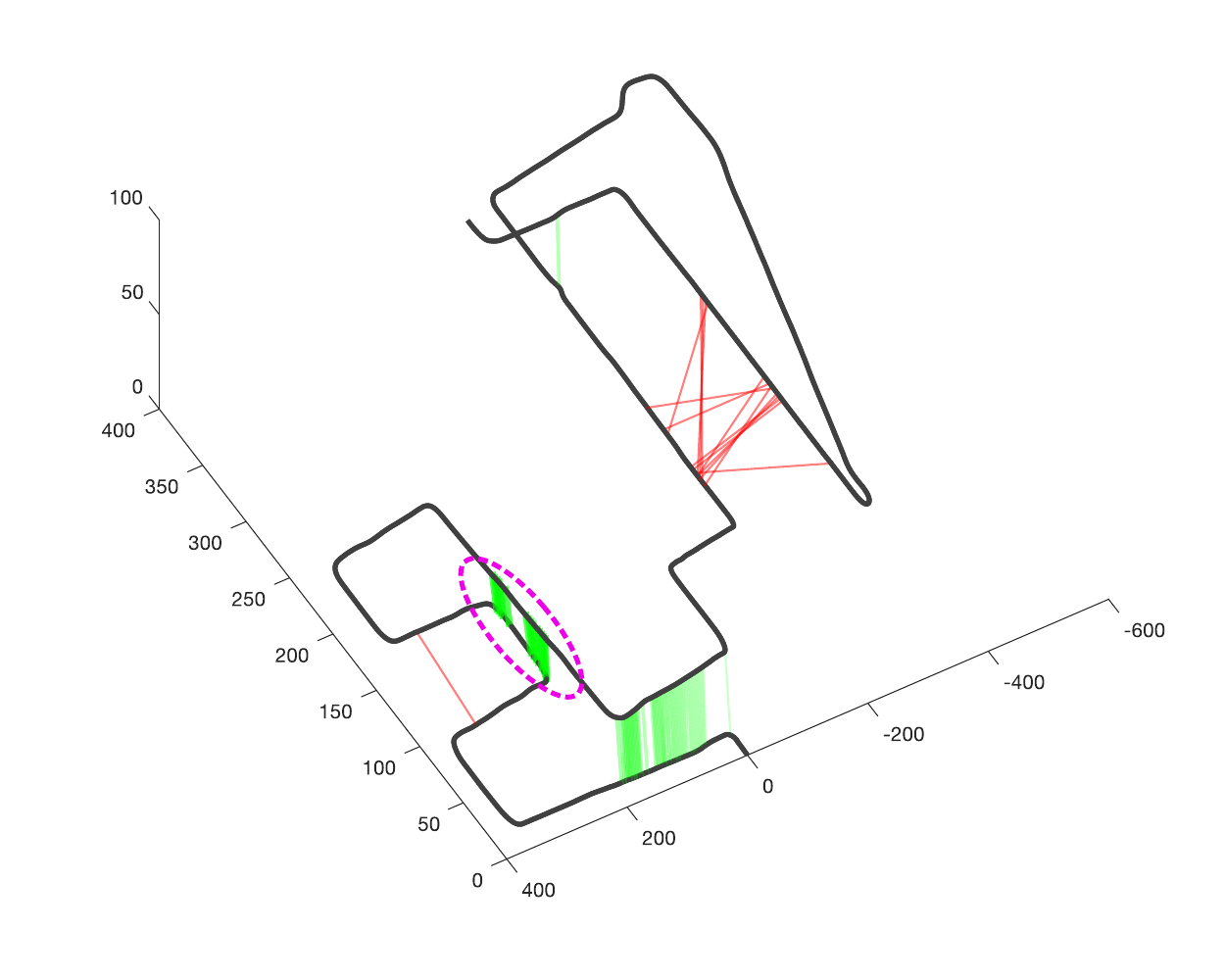}}
\subfloat[LiDAR ORB]{\includegraphics[width=0.2\paperwidth]{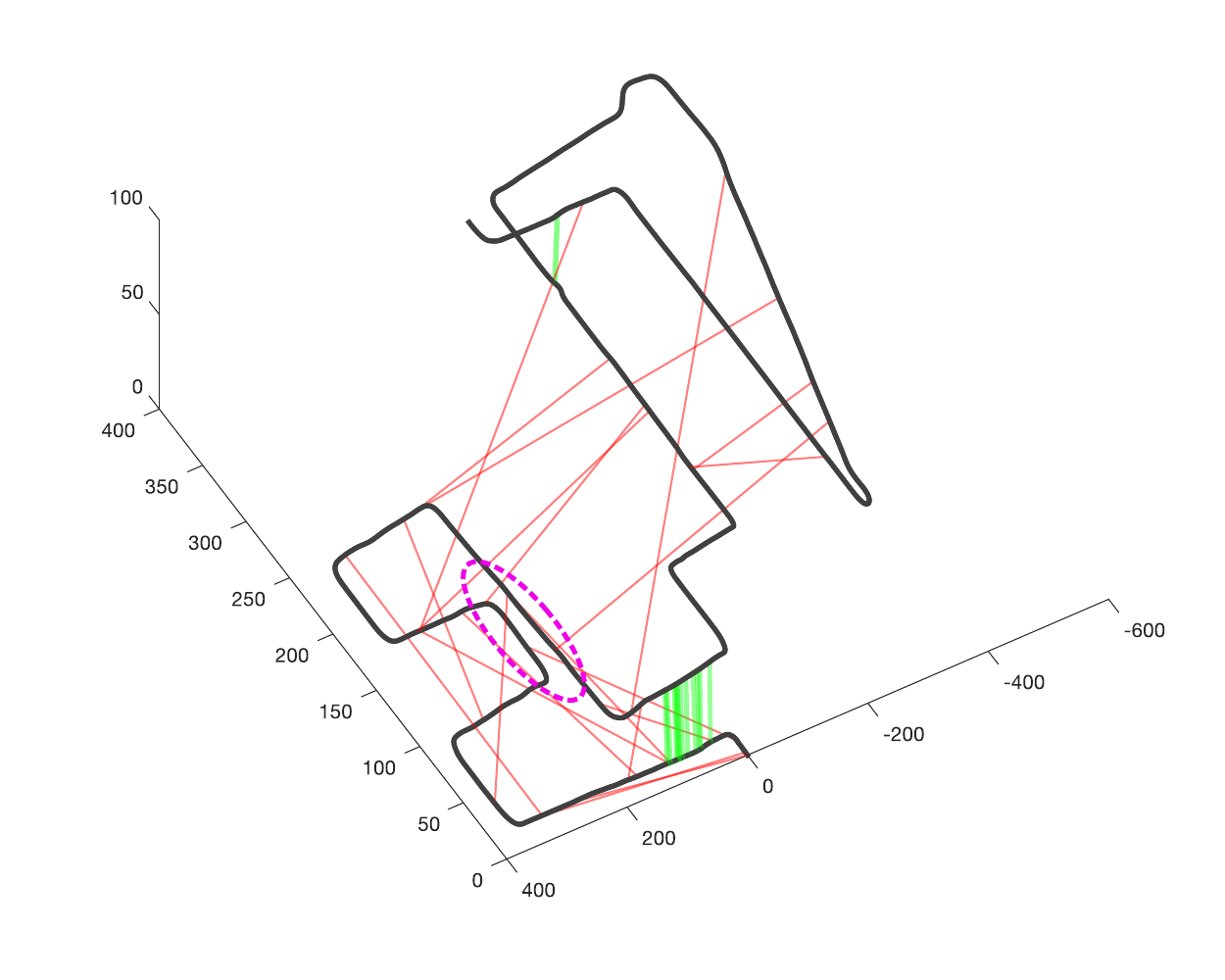}}
\caption{Place recognition result on KITTI dataset.}
\label{fig_result_kitti}
\end{figure}

\begin{figure}
\begin{centering}
\subfloat[FreSCo (ours)]{\includegraphics[width=0.2\paperwidth]{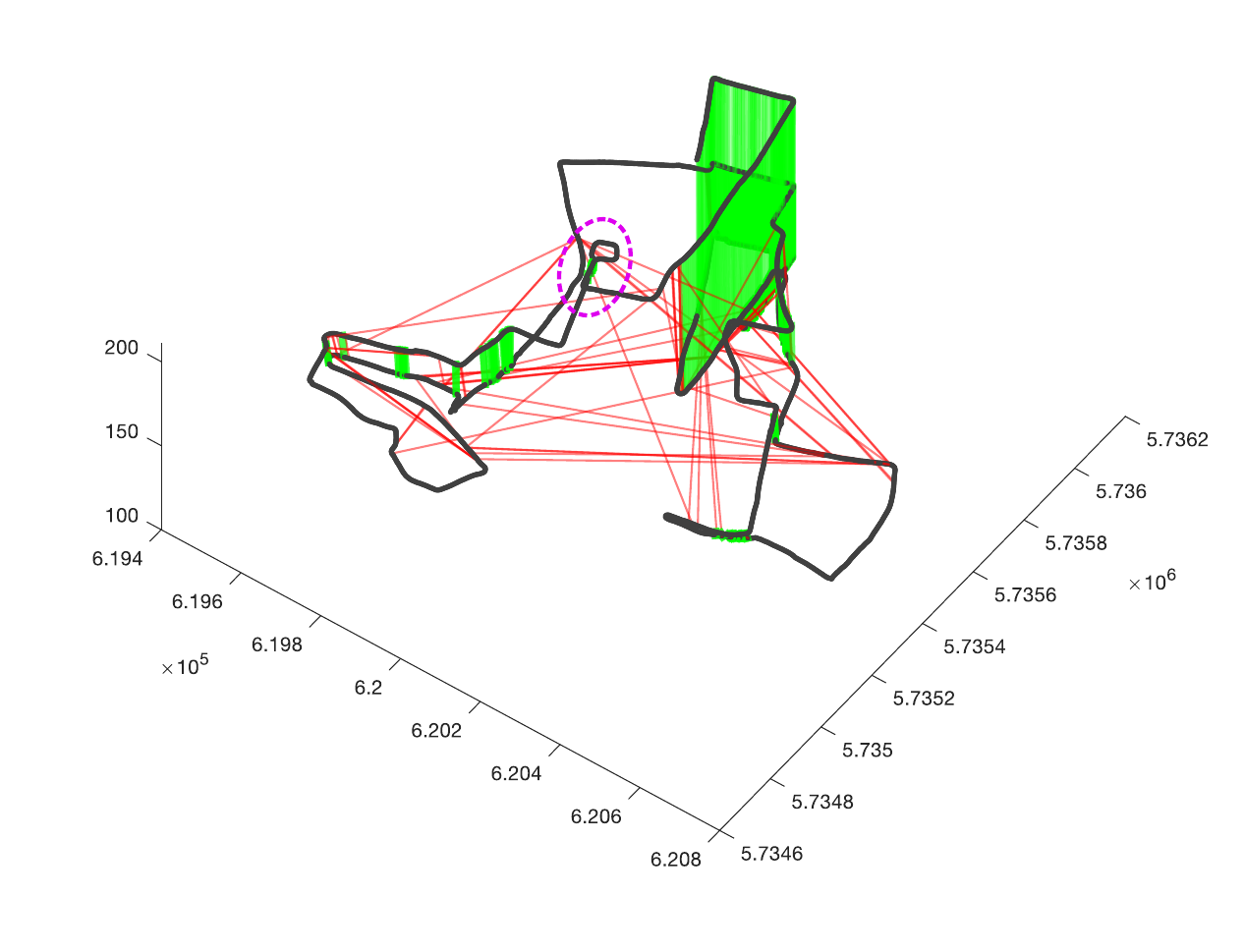}}
\subfloat[Scan Context]{\includegraphics[width=0.2\paperwidth]{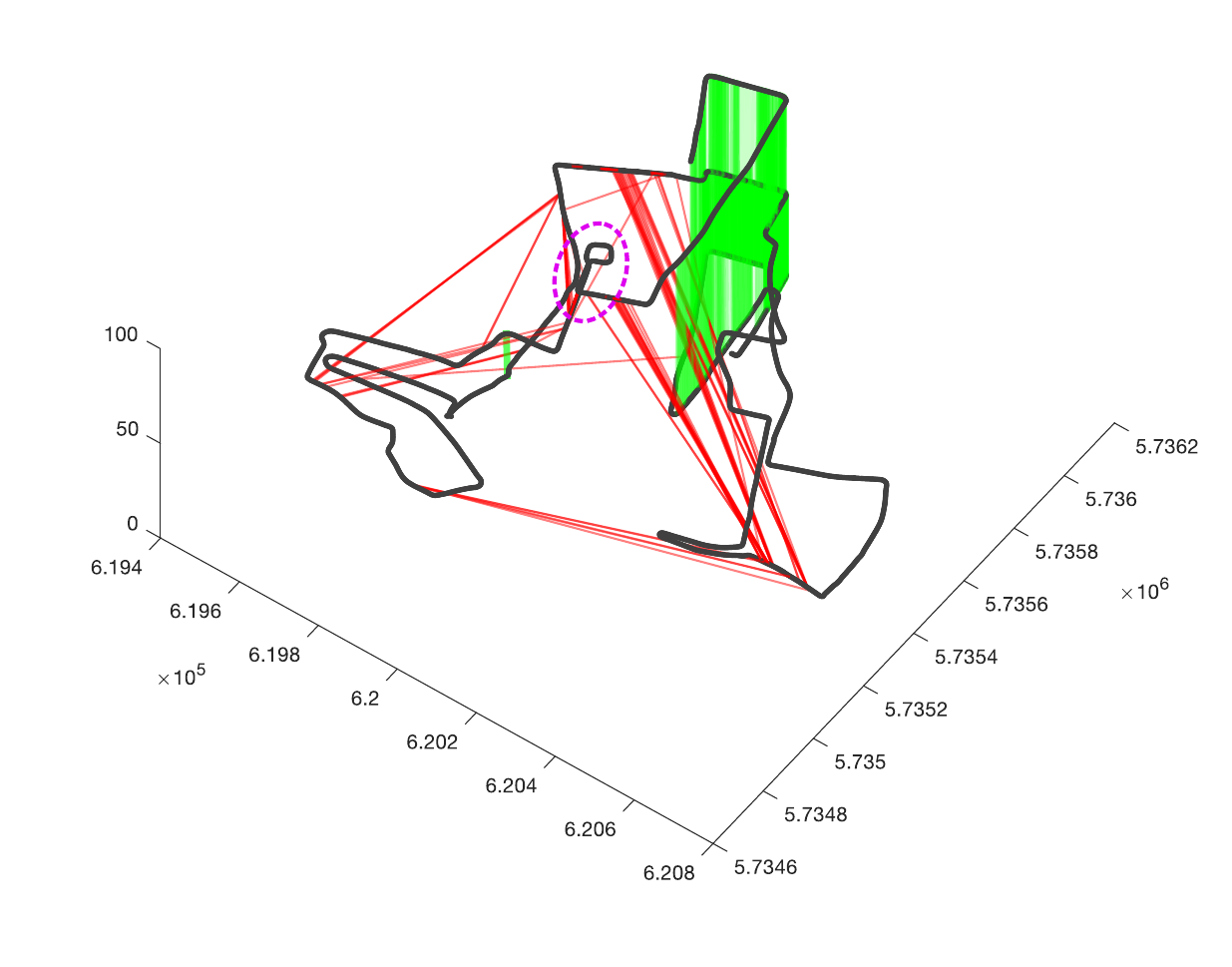}}
\par\end{centering}
\begin{centering}
\subfloat[LiDAR Iris]{\includegraphics[width=0.2\paperwidth]{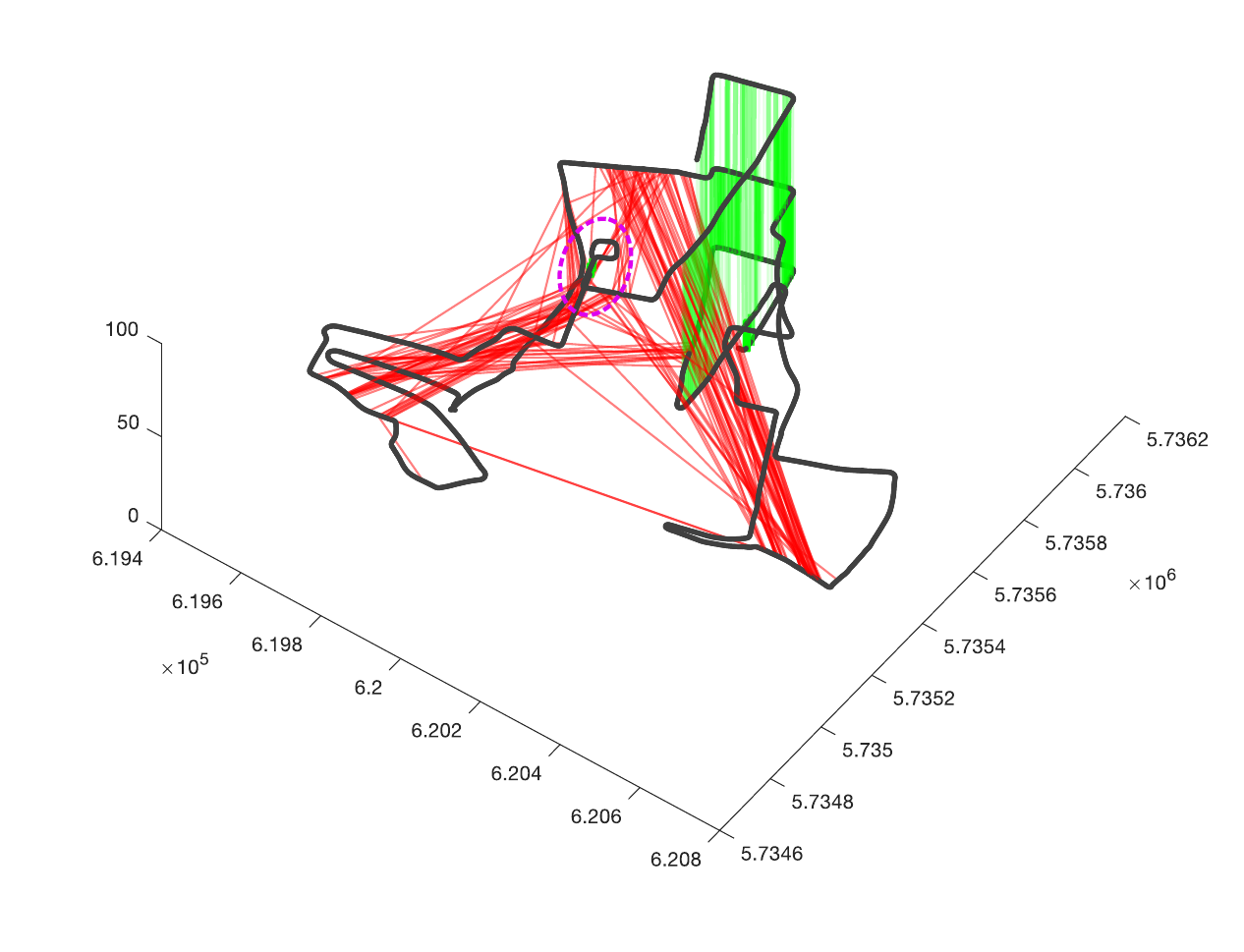}}
\subfloat[LiDAR ORB]{\includegraphics[width=0.2\paperwidth]{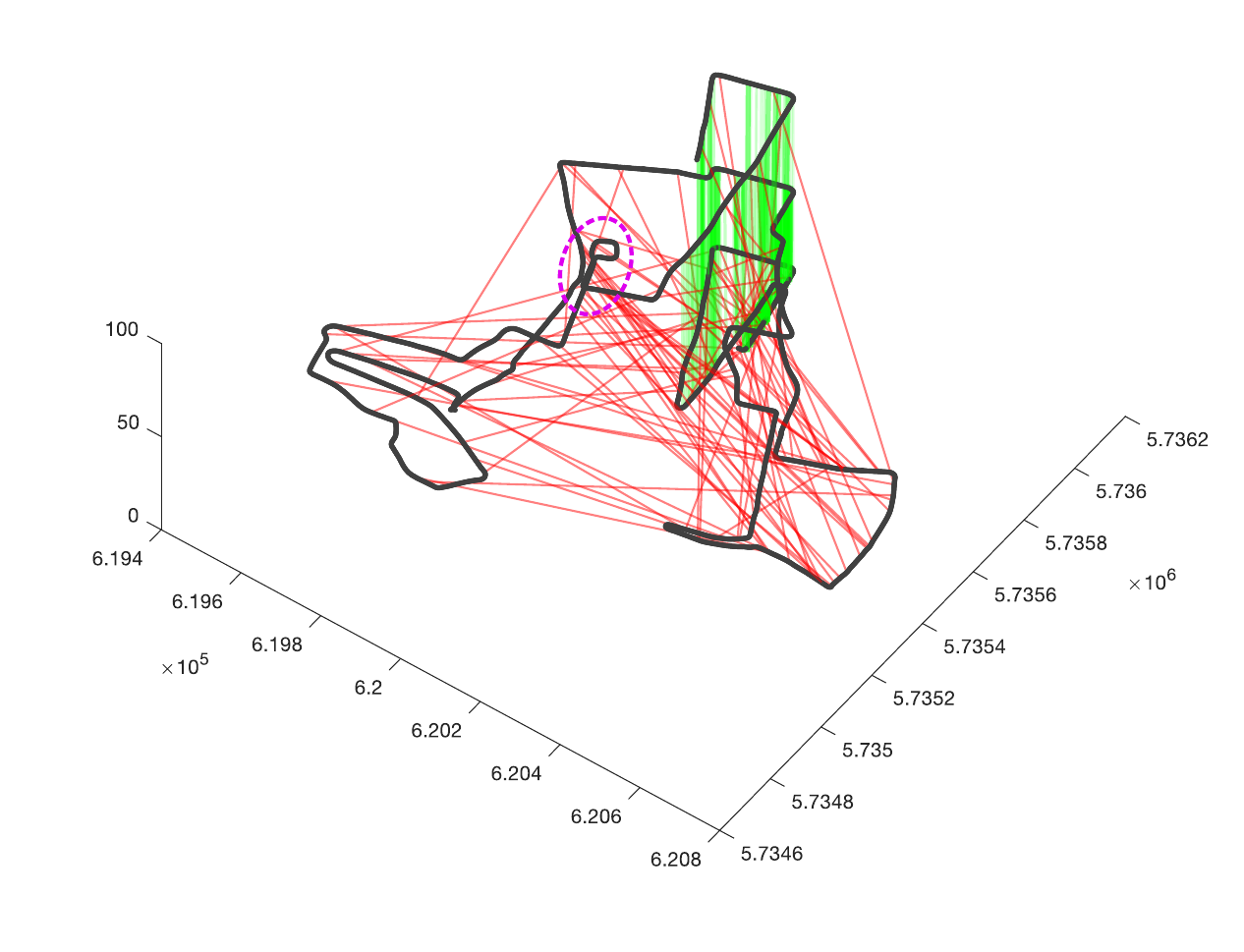}}
\par\end{centering}
\caption{Place recognition result on Oxford dataset.}
\label{fig_result_oxford}
\end{figure}

Precision-recall curves in Fig.~\ref{fig_pr_curves} show the performance of place retrieval. 
LiDAR ORB exhibits the poorest performance among all approaches.
It solely relies on the feature points extracted from LiDAR intensity images. 
However, due to the sparsity nature of the 32-beam or 64-beam LiDAR measurements, the features are not well repeatable. 
Its performance degenerates significantly on KITTI and Oxford datasets, as the variation of intensity readings between lasers from Velodyne's LiDAR is reported higher than that of Ouster's. 
Our method outperforms other contestants over all datasets. 
Detailed results are shown in Fig.~\ref{fig_result_mulran}, Fig.~\ref{fig_result_kitti}, Fig.~\ref{fig_result_oxford} and
Table~\ref{tab_comparison}. 
Black curves are the trajectories of each sequence, elevated on the $z$-axis over time for the convenience of visualization. 
Green and red line segments correspond to the true positive and false positive matches, respectively. 
All results are obtained with the threshold configuration at the highest F1 score for each approach. 

\begin{figure}[t]
\centering{}
\subfloat[]{\includegraphics[width=0.32\columnwidth]{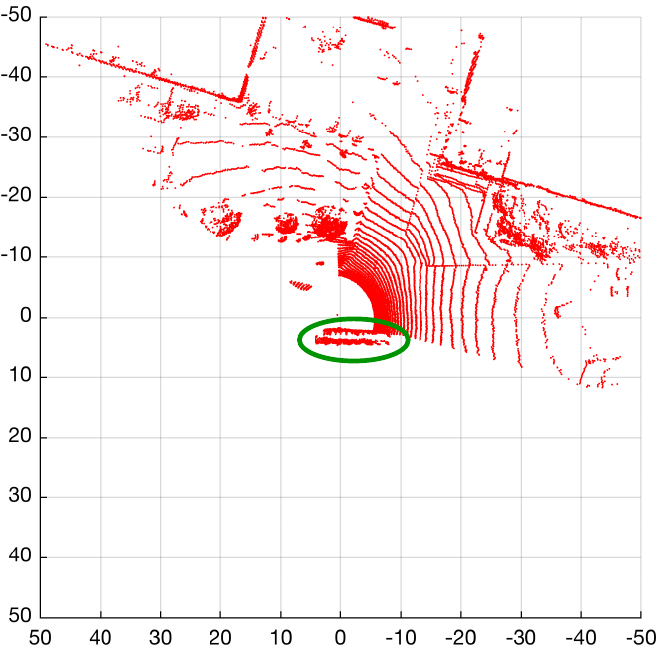}}
\subfloat[]{\includegraphics[width=0.32\columnwidth]{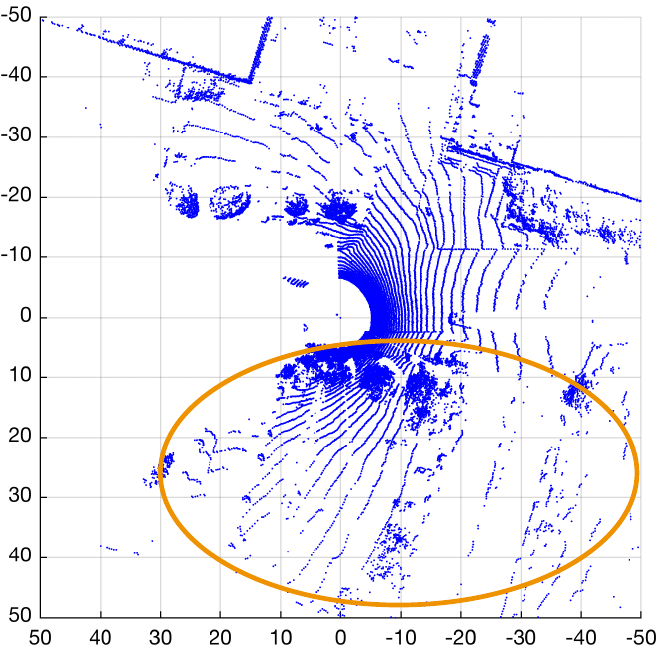}}
\subfloat[]{\includegraphics[width=0.32\columnwidth]{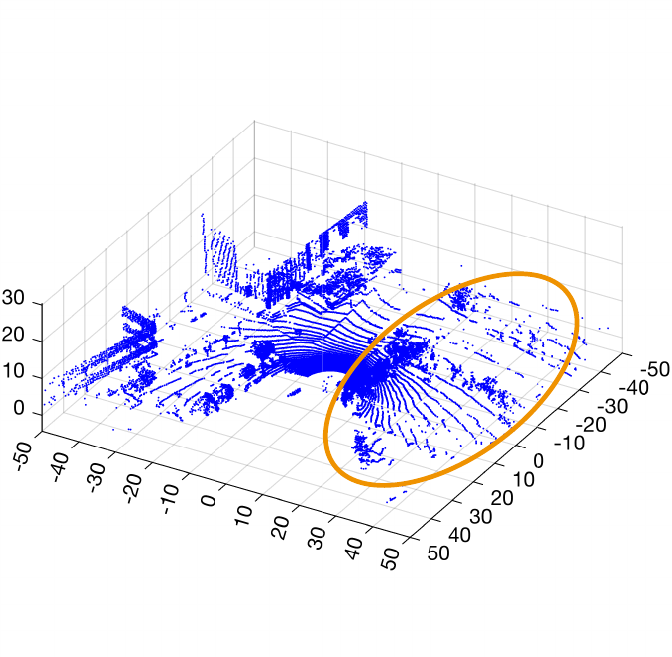}}\caption{\label{fig_dynamic_in_mulran}Occurrence of dynamic obstacle during
revisit. (a) and (b) are the top views of the point cloud of $6590^{{\rm th}}$
and $4179^{{\rm th}}$ frame from MulRan \texttt{KAIST 03}, respectively.
(c) is the oblique view of the point cloud shown in (b).}
\end{figure}

When evaluated on the MulRan dataset, Scan Context and FreSCo reveal similar results as shown in Fig.~\ref{fig_result_mulran}. 
This is because \texttt{KAIST 03} only has revisits of the same direction, and little translation occurs during revisits, which bring little challenge. 
Despite that, it should be noted that difference exists in the magenta dashed circles. 
This piece of trajectory encounters a moving bus as the dynamic obstacle to the LiDAR, which is shown in Fig.~\ref{fig_dynamic_in_mulran}.
Occlusion circled in orange is caused by the bus circled in green.
Although the occluded area does not contain much distinctive information, Scan Context and LiDAR Iris still failed to find correct matches in this area. 
This is due to the fact that occlusion jeopardizes the integrity of the spacial global descriptor in the occluded angle range, and Scan Context and LiDAR Iris value the signature on each orientation equally. 
In FreSCo, more signatures are extracted from regions with richer frequency components, and they are not valued equally on each spacial orientation, making it less sensitive to occlusions. 

\begin{figure}
\begin{centering}
\subfloat[]{
\includegraphics[width=0.48\columnwidth]{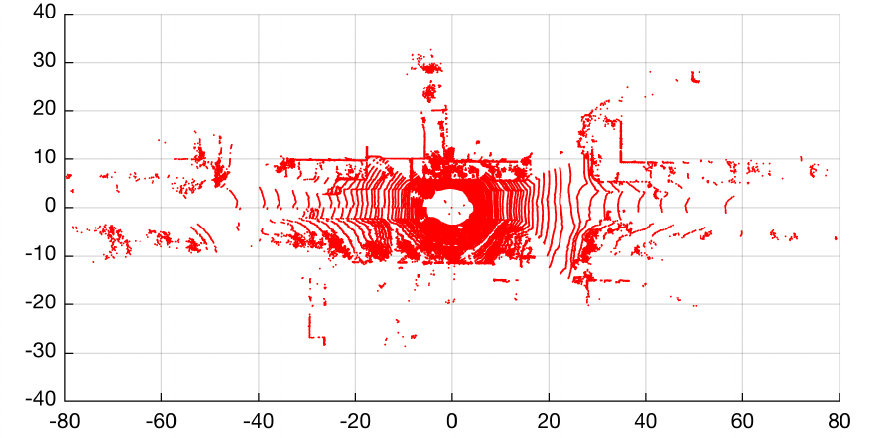}
}
\subfloat[]{
\includegraphics[width=0.48\columnwidth]{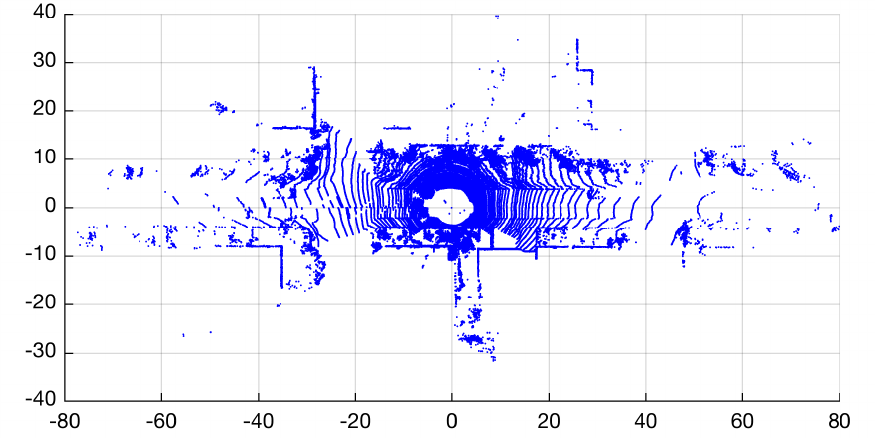}
}
\caption{A scene with low structural diversity
during revisit. (a) and (b) are the top views of the point cloud of
$1466^{{\rm th}}$ and $742^{{\rm nd}}$ frame from KITTI \texttt{08},
respectively. }
\label{fig_low_diversity_in_kitti}
\par
\end{centering}
\end{figure}

The results reported on the KITTI dataset are shown in Fig.~\ref{fig_result_kitti}.
Notice that all approaches except for our FreSCo miss matches notably in the circled area. 
Revisits within this area are in the reversed direction and have strafe translation of about $1.5{\rm m}$, and the scenes have low structural diversity. 
One of them is shown in Fig.~\ref{fig_low_diversity_in_kitti}.
Scan Context and LiDAR Iris leverage global descriptors generated from egocentric polar coordinates, which is not translation-invariant.
Although they use coarse discretization on the radial axis to adapt for small translations, this compromises the extraction of fine details, resulting in deteriorative performances
in scenes with low structural diversities. 
By contrast, our FreSCo achieves both translation invariance and the ability to capture fine-grained details. 
Hence successful matches on revisits with even large translations can be observed on both ends of this piece of trajectory. 

The Oxford dataset poses the most challenges, including more occurrences of dynamic objects, a higher ratio of scenes with similar patterns, larger translations during revisits, and also constant occlusion caused by other sensors mounted nearby. 
The results reported are shown in Fig.~\ref{fig_result_oxford}. 
Since the vehicle drives in different lanes when reversed revisits happen, strafe translation in such cases reaches $3{\rm m}$ and above. 
And corridor-like scenes can be observed in the circled area. 
In spite of the complexity discussed above, our FreSCo still outperforms other approaches by a large margin.

\subsection{Pose Estimation}

\begin{table}[t]
	\centering{}%
	\caption{\label{tab_accuracy}Accuracy of first stage pose estimation and \newline overall success rate.}
	\begin{tabular}{>{\raggedright}p{2cm}ccccc}
		\toprule 
		\multirow{2}{2cm}{Approach} & \multicolumn{2}{>{\centering}m{1.6cm}}{RTE (m)} & \multicolumn{2}{>{\centering}m{1.6cm}}{RRE (deg)} & \multirow{2}{*}{Overall SR (\%)}\tabularnewline
		\cmidrule{2-5} \cmidrule{3-5} \cmidrule{4-5} \cmidrule{5-5} 
		& Mean & Std. & Mean & Std. & \tabularnewline
		\midrule 
		Scan Context & - & - & - & - & 56.41\tabularnewline
		\midrule 
		\textbf{FreSCo (ours)} & \textbf{0.23} & \textbf{0.15} & \textbf{0.37} & \textbf{0.32} & \textbf{96.36}\tabularnewline
		\bottomrule
	\end{tabular}
\end{table}

The two-stage pose estimation process gives the relative pose after
place retrieval. Since the 3D ICP refinement for the relative pose
is optional, we only evaluated the accuracy of the relative pose obtained
in the first stage. Relative translation error (RTE) and relative
rotation error (RRE) are used to quantify the accuracy. Only true
positive matches are used for this evaluation. Among methods for comparison,
only Scan Context provides pose estimation and it is a one-stage method
by directly running 3D ICP with an initial guess for the relative
rotation. Therefore, we only compare the overall success rate (SR)
for them. A result of pose estimation is regarded successful when
the MSE between two aligned 3D point clouds is less than $1.5 m^2$.

Experiments are conducted on KITTI \texttt{08} sequence, and the results
are shown in Table \ref{tab_accuracy}. The SR metric indicates that
the pose estimation method in FreSCo is far more robust than that
of Scan Context. This is due to two reasons: 1) Some of the true positive
matches of Scan Context obtained in the place retrieval phase have
wrong initial guesses for the relative rotation. 2) 3D ICP is very
sensitive to initial guess, thus 1-DoF initial guess is not enough
to guarantee a good pose estimation result. By contrast, FreSCo provides
better initial guess for relative rotation, and the first stage of
pose estimation exploits 2D NICP, which is less sensitive to initial
guess.

\subsection{Runtime Evaluation}

\begin{table}[t]
	\begin{centering}
		\caption{\label{tab_runtime}Comparison on runtime of each phase. }
		\begin{tabular}{l>{\centering}p{1.2cm}>{\centering}p{1.2cm}>{\centering}p{1.2cm}>{\centering}p{1.2cm}}
			\toprule 
			& Descriptor & Place & 1\textsuperscript{st} Stage & 2\textsuperscript{nd} Stage\tabularnewline
			Approach & Gen. & Retrieval & Pose Est. & Pose Est.\tabularnewline
			& (ms) & (ms) & (ms) & (ms)\tabularnewline
			\midrule 
			Scan Context & 4 & 8 & - & 1020\tabularnewline
			\midrule 
			LiDAR Iris & 10 & 190 & - & -\tabularnewline
			\midrule 
			LiDAR ORB & 45 & 94 & - & -\tabularnewline
			\midrule 
			\textbf{FreSCo (ours)} & \textbf{26} & \textbf{7} & \textbf{143} & \textbf{593}\textsuperscript{*}\tabularnewline
			\bottomrule
		\end{tabular}
		\par\end{centering}
\end{table}

We implemented the preprocessing, BEV image generation and pose estimation
in C++, and the FreSCo main framework in MATLAB. All experiments are
conducted on an Intel NUC 8 platform with an Intel i7-8559U processor
at $2.70{\rm GHz}$ and a memory of $16{\rm GB}$. The average runtime
for each frame evaluated on KITTI \texttt{08} are shown in Table \ref{tab_runtime}.
Note that our FreSCo provides good relative pose at the first
stage of pose estimation, the second stage computation can
be omitted if the need for accuracy is satisfied.

\section{Conclusion}

In this paper, we presented FreSCo, a novel global descriptor introduced in the frequency domain.
Specifically, FreSCo exploits Fourier Transform on the bird's eye view image in the Cartesian coordinate system and achieves both translation and rotation invariance. 
Compared to existing methods, ours showed better performance in place retrieval over various scenes. 
We also introduced a fast pose estimation method, which leverages flattened 2D point clouds and achieves good accuracy at a low computational cost. 
FreSCo is an application-oriented approach that does not depend on special computing platforms to achieve satisfactory speeds. 
We hope that it can be deployed on edge devices in the future to lay the foundation for high-level tasks such as autonomous driving.

\bibliographystyle{IEEEtran}
\bibliography{bibs_full_name}

\end{document}